\pdfoutput=1

\documentclass[11pt]{article}

\usepackage[preprint]{acl}

\usepackage{times}
\usepackage{latexsym}

\usepackage[T1]{fontenc}

\usepackage[utf8]{inputenc}

\usepackage{microtype}

\usepackage{inconsolata}

\usepackage{graphicx}
\usepackage{caption}
\usepackage{subcaption}

\usepackage{tcolorbox}
\newtcbox{\code}[1][]{
  colback=gray!10!white,
  colframe=gray!20!white,
  boxrule=1pt,
  left=0mm,right=0mm,top=0mm,bottom=0mm,
  box align=base,
  nobeforeafter,
  fontupper=\ttfamily
}
\usepackage{listings}
\usepackage{xcolor}

\definecolor{codegreen}{rgb}{0,0.6,0}
\definecolor{codegray}{rgb}{0.5,0.5,0.5}
\definecolor{codepurple}{rgb}{0.58,0,0.82}
\definecolor{backcolour}{rgb}{0.95,0.95,0.92}

\lstdefinestyle{mystyle}{
    backgroundcolor=\color{backcolour},   
    commentstyle=\color{codegreen},
    keywordstyle=\color{magenta},
    numberstyle=\tiny\color{codegray},
    stringstyle=\color{codepurple},
    basicstyle=\ttfamily\footnotesize,
    breakatwhitespace=false,         
    breaklines=true,                 
    captionpos=b,                    
    keepspaces=true,                 
    numbers=left,                    
    numbersep=5pt,                  
    showspaces=false,                
    showstringspaces=false,
    showtabs=false,                  
    tabsize=2
}

\usepackage{amsmath,amssymb,bm}
\usepackage{url}
\usepackage {diagbox}
\usepackage[ruled, linesnumbered, titlenumbered]{algorithm2e}
\usepackage{bbm}
\usepackage{multirow}
\usepackage {booktabs}

\title{Tonguescape: Exploring Language Models Understanding of \\Vowel Articulation}

\author{
  \textbf{Haruki Sakajo},
  \textbf{Yusuke Sakai},
  \textbf{Hidetaka Kamigaito},
  \textbf{Taro Watanabe}
\\
  Nara Institute of Science and Technology (NAIST)
\\
\texttt{sakajo.haruki.sd9@naist.ac.jp}\\
\texttt{\{sakai.yusuke.sr9, kamigaito.h, taro\}@is.naist.jp}
}

\begin{document}
\maketitle
\begin{abstract}
Vowels are primarily characterized by tongue position.
Humans have discovered these features of vowel articulation through their own experience and explicit objective observation such as using MRI.
With this knowledge and our experience, we can explain and understand the relationship between tongue positions and vowels, and this knowledge is helpful for language learners to learn pronunciation.
Since language models (LMs) are trained on a large amount of data that includes linguistic and medical fields, our preliminary studies indicate that an LM is able to explain the pronunciation mechanisms of vowels.
However, it is unclear whether multi-modal LMs, such as vision LMs, align textual information with visual information.
One question arises: do LMs associate real tongue positions with vowel articulation?
In this study, we created video and image datasets from the existing real-time MRI dataset and investigated whether LMs can understand vowel articulation based on tongue positions using vision-based information.
Our findings suggest that LMs exhibit potential for understanding vowels and tongue positions when reference examples are provided while they have difficulties without them.
Our code for dataset building is available on GitHub~\footnote{\url{https://github.com/sj-h4/tonguescape-builder}}.

\end{abstract}

\section{Introduction}
In phonetics, vowels are distinguished and described by focusing on tongue positions and lip shape.
Beginning with \citet{jonesEverymansEnglishPronouncing1917}, humans have explained vowels based on tongue positions during articulation.
Human speakers are aware of speech mechanisms with training through introspection of experience and observations of visual information (e.g., MRI).
For example, when pronouncing the English word ``image,'' speakers can perceive and explain that the initial sound is produced by positioning the tongue forward and high in the mouth.
Moreover, speakers can relatively recognize vowels, using a certain vowel as a reference point.
This is the method \citet{jonesEverymansEnglishPronouncing1917} employed to introduce the cardinal vowels, demonstrating their grasp of tongue positions through introspection and objective observations, and to link this understanding to the concept of vowel articulation.
This knowledge helps explain to language learners how to pronounce and describe linguistic phenomena.

\begin{figure}[t]
\centering
  \includegraphics[width=\columnwidth]{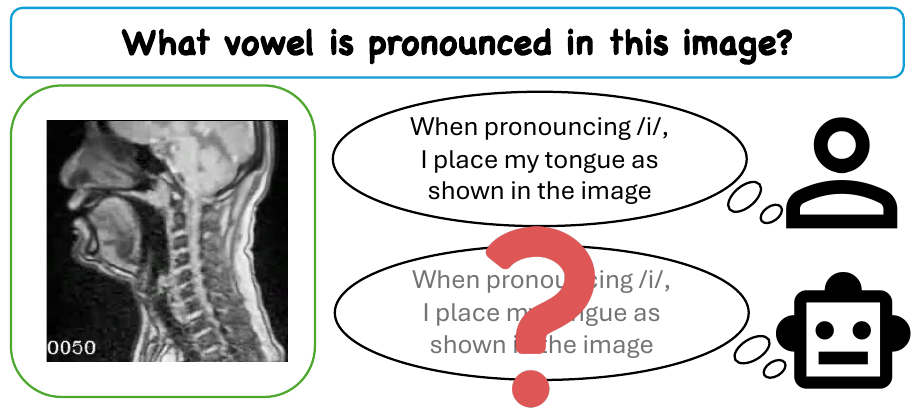}
  \caption{An illustration of our research question.}
  \label{fig:introduction}
\end{figure}

Language Models (LMs) are trained on a large amount of data that includes linguistic and medical fields. 
Our preliminary studies showed that LMs know vowel pronunciation and the correlation between vowels and tongue positions as textual knowledge (see Section \ref{sec:backgroung_related_works}).
To determine if LMs comprehend articulation relative to the articulatory organs, multi-modal information is essential.
Multi-modal LMs capture not only textual information, but also images, videos, and audio~\citep{zhou-etal-2023-rome, li2024llavanext-interleave, geminiteamGeminiUnlockingMultimodal2024}. Moreover, their application is expanding to more specialized fields, such as clinical tasks involving the detection and explanation of diseases from clinical images such as CT and MRI~\citep{yanWorseRandomEmbarrassingly2024, pal-sankarasubbu-2024-gemini}.
However, it is known that the alignment among modalities, such as images and text, is often weak in multi-modal LMs, and it remains unclear whether these models truly understand such interactions~\citep{caoUnderstandingAttentionVisionLanguage2022, kawaharazukaRealWorldRobotApplications2024, hayashi-etal-2024-towards}.

\begin{figure*}[t]
    \centering
    \includegraphics[width=\textwidth]{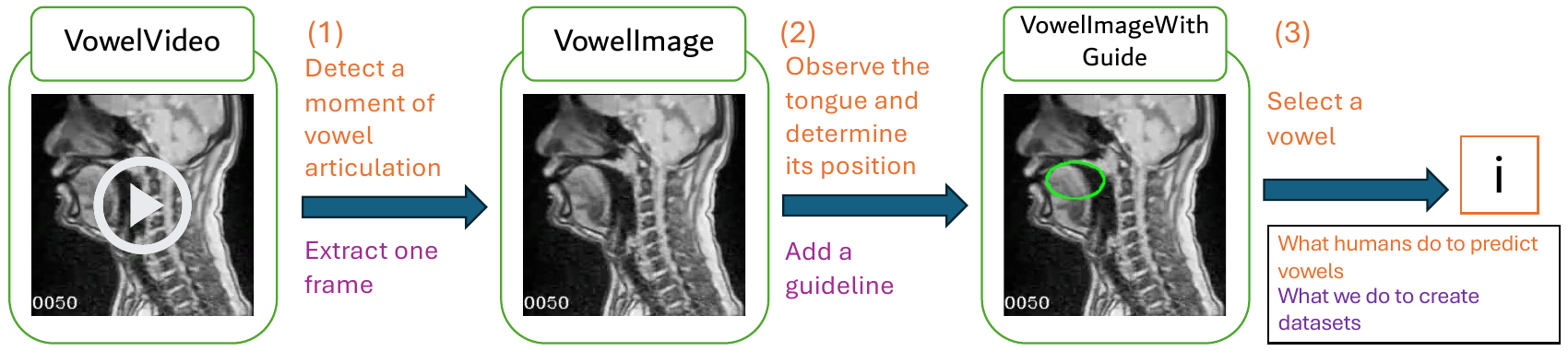}
    \caption{Examples of our dataset and what human speakers do when predicting vowels from real-time MRI and what we did when we constructed our dataset to replicate it. VowelVideo simulates the process of humans predicting vowels from entire pronunciation. VowelImage simulates the observation of a moment of articulation. VowelImageWithGuide simulates the examination of tongue positions within the oral cavity.}
    \label{fig:dataset_example}
\end{figure*}

\emph{Do LMs understand tongue positions through knowledge and objective observation and asso- ciate them with the concept of vowel articulation?}
We ask this research question as illustrated in Figure~\ref{fig:introduction}.
If LMs can align the visual information of articulatory processes with textual descriptions of phonetics, they could contribute to further analysis of linguistic phenomena, such as vowel harmony, and assist in pronunciation education. They could also aid in making speech understanding more grounded in human physiology, contributing to a more embodied multi-modal understanding.

In this study, we curated and annotated videos and images of tongue positions from real-time MRI of articulatory movements as illustrated in Figure~\ref{fig:dataset_example}.
We examined whether LMs can truly understand vowel articulation based on tongue positions in a visual/video question-answering format.
We found that some models seem to have some ability to predict tongue positions and vowels when provided with images of each vowel in a few-shot setting, while the other models have difficulties in recognizing them either in zero-shot, few-shot, and fine-tuning settings.

\section{Backgound and Related Works}
\label{sec:backgroung_related_works}

\subsection{Tongue and Vowels in Linguistics}
In phonetics, vowels are described by the height and blackness of the body of the tongue in the oral cavity.
The tongue positions are measured by using relative positions in the vowel categories rather than absolute positions~\citep{jonesEverymansEnglishPronouncing1917, knightCambridgeHandbookPhonetics2021a}.
These characteristics describe several linguistic phenomena~\citep{knightCambridgeHandbookPhonetics2021a} and are also adopted in the International Phonetic Alphabet (IPA)~\citep{associationHandbookInternationalPhonetic1999}.
For example, when pronouncing the vowel /i/ in Figure~\ref{fig:introduction}, the tongue is moved forward and raised higher, creating a narrow space in the anterior region of the oral cavity.

\begin{figure}[t]
    \centering
    \includegraphics[width=0.6\columnwidth]{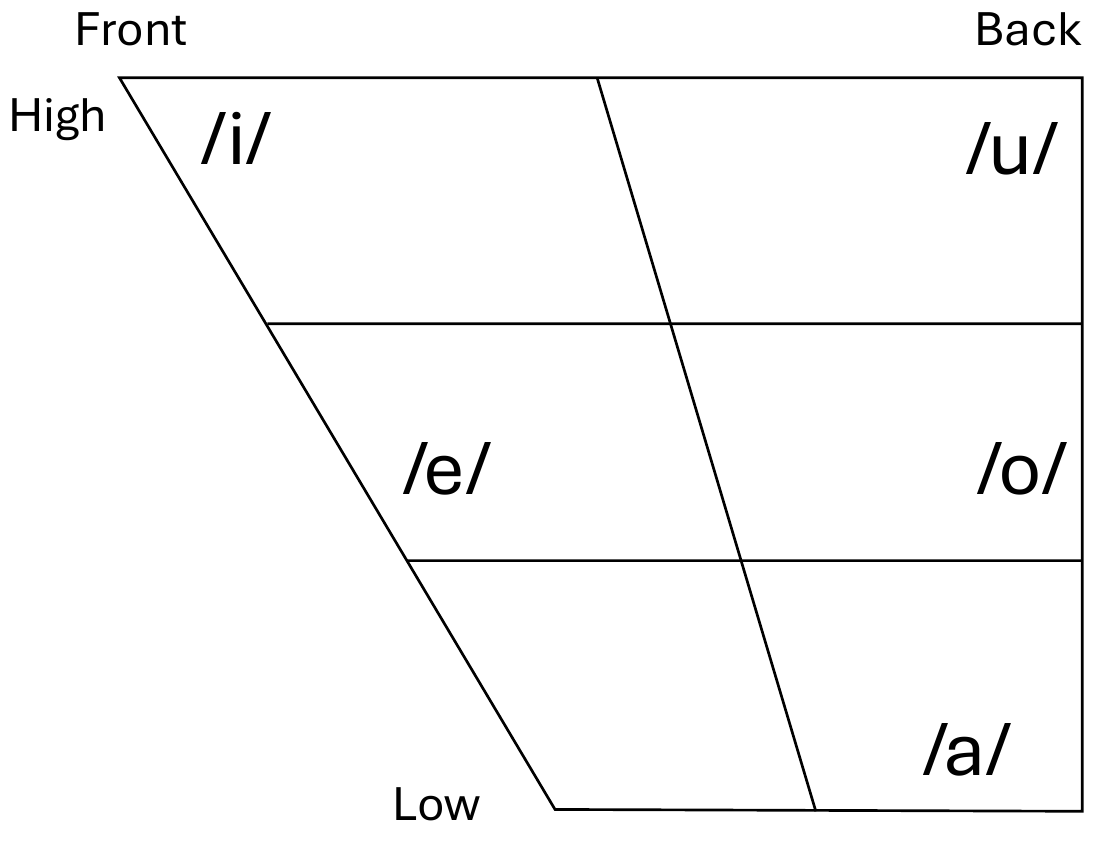}
    \caption{An example of vowel chart in five-vowel system.}
    \label{fig:japanese_vowel}
\end{figure}

The five-vowel system (/a/, /i/, /u/, /e/, /o/) is one of the most common vowel inventories in world languages ~\citep{kubozonoHandbookJapanesePhonetics2015}, and this size of a vowel inventory accounts for more than half of languages~\citep{wals-vowel-inventories}.
These vowels are characterized by tongue height and backness: high-mid-low and front-back~\footnote{Vowel height can be explained as closeness. In this study, we use high-mid-low because preliminary studies show that LMs explain vowels using height.}.
The tongue positions of each vowel are /a/~(low-back), /i/~(high-front), /u/~(high-back), /e/~(mid-front), and /o/~(mid-back), as shown in Figure~\ref{fig:japanese_vowel}.
The tongue position for the vowel /o/ is intermediate between /a/ and /u/.

Speakers perceive and understand tongue positions during speech, allowing them to pronounce a vowel when instructed to say it between /a/ and /e/ with training.
This also enables us to explain linguistic phenomena related to tongue positions, such as vowel harmony~\cite{Bybee_2015}.
However, studying these phenomena sometimes requires introducing new but consistent principles, as introduced by \citet{ko-2012-tongue}, \citet{joseph-2018-historical}, and \citet{WangManZhouYuXianDaiFangYanniokeruMuYinDiaoHe2020}.
If LMs can understand the articulation, they can provide researchers with valuable insights into these underlying principles.

\begin{table*}[!t]
    \centering
    \small
    \resizebox{\linewidth}{!}{%
    \setlength{\tabcolsep}{2pt}
    \begin{tabular}{@{}cp{14cm}@{}}
    \toprule
        Query & There are five vowels: /a/ /i/ /u/ /e/ /o/. Please explain briefly how to pronounce the given vowel. : i  \\
    \midrule
        Answer & The vowel /i/ is pronounced as a high front unrounded vowel. To produce this sound, follow these steps:\newline \ \newline 1. Position your tongue high in your mouth, close to the roof but not touching it.\newline 2. Move your tongue towards the front of your mouth.\newline 3. Keep your lips unrounded and slightly spread, as if you are smiling.\newline 4. Allow your vocal cords to vibrate as you produce the sound.\newline \ \newline This vowel sound is similar to the ``ee'' in the English word ``see.'' \\
    \bottomrule
    \end{tabular}
    }
    \caption{An example of an explanation of a vowel /i/ by GPT-4o. LMs are able to explain how to pronounce a vowel, focusing on tongue positions and lip shape. The explanation is correct linguistically, and speakers are able to pronounce the vowel by following this instruction.}
    \label{tab:explanation_vowel_tonge}
\end{table*}

\begin{table*}[!t]
    \centering
    \small
    \setlength{\tabcolsep}{2pt}
    \resizebox{\linewidth}{!}{%
    \begin{tabular}{@{}cp{14cm}@{}}
    \toprule
        Query & There are five vowels: /a/ /i/ /u/ /e/ /o/. You can predict which vowel is pronounced by reading the position of the tongue. First, read the tongue height and backness. Use this information to predict which vowel is pronounced in the given information. Respond with either a, i, u, e, or o as your answer. : high, front  \\
    \midrule
        Answer & i\\
    \bottomrule
    \end{tabular}
    }
    \caption{An example of a prediction of a vowel from the given tongue position by GPT-4o. LMs appear to have the capability of predicting vowels from textual information by using their textual knowledge.}
    \label{tab:prediction_vowel_tonge}
\end{table*}

\subsection{Tongue and Vowels in Language Models}

Some studies focus on vowels and tongue-related information.
\citet{steuer-etal-2023-information} found that phoneme-level LMs capture vowel harmony, a linguistic phenomenon related to tongue position.
\citet{cotterell-eisner-2017-probabilistic} predicted vowel inventories by introducing models based on stochastic point processes.
They used formant information to identify vowels influenced by tongue position.
\citet{cotterell-eisner-2018-deep} also used formant information to describe phonetic-related linguistic typology by a generative probability model.
However, since these studies focus on textual information, it remains unclear whether multi-modal LMs can achieve an embodied understanding of the relationship between tongue positions and vowels with aligned textual and visual information.

\subsection{Application of Understanding Tongue Positions}
Tongue positions are also used to improve speech synthesis systems.
\citet{abeysingheVisualisingModelTraining2022} found that the vowel space of a speech synthesis model changes during fine-tuning and that it can be visualized.
\citet{wuDeepSpeechSynthesis2023} introduced a speech synthesis method that uses MRI-based features and demonstrates that MRI provides useful features for synthesizing speech.
If LMs possess the capability to comprehend the relationship between tongue position and articulation, it could further advance these studies, contributing to the synthesis of fluent or intentionally disfluent speech as human speech.

\subsection{Preliminary Studies}
We investigated whether LMs know the relationship between vowels and tongue positions using GPT-4o~\citep{gpt4o}.
Table~\ref{tab:explanation_vowel_tonge} demonstrates that they should be able to coherently explain the relationship between tongue position and vowel articulation like human speakers.
Furthermore, Table~\ref{tab:prediction_vowel_tonge} shows that they are also capable of predicting a vowel from the given tongue position.
We used each tongue position of the five vowels as a query, and GPT-4o answered the correct vowel.
Therefore, these preliminary studies show that LMs know how to pronounce vowels.

\section{Dataset: Tonguescape}
\label{sec:dataset}

When predicting vowels from a real-time MRI, speakers will (1) detect the moment of vowel articulation in the video, (2) observe the tongue and determine its position, and then (3) select a vowel. We propose a QA dataset for vowel prediction from real-time MRI recordings of tongue movements during vowel articulation comprising three steps, with each step corresponding to these stages of human perception as shown in Figure~\ref{fig:dataset_example}.
We curated and annotated videos and images of tongue positions from the Real-time MRI Articulatory Movement Database (rtMRIDB)~\citep{maekawaRealtimeMRIArticulatory2022}, which comprises real-time MRI recordings of articulatory movements of Japanese phonemes.
In this study, we focused on the Japanese five-vowel system (/a/, /i/, /u/, /e/, /o/) where each vowel has distinct articulatory features.
Details of our dataset are described in Appendix~\ref{sec:detail-rtmridb}.


\subsection{Real-time MRI Articulatory Movement Datasets}

The Real-time MRI Articulatory Movement Database~(rtMRIDB)~\citep{maekawaRealtimeMRIArticulatory2022} is the dataset that contains videos recording the articulations of Japanese phonemes.
The dataset consists of utterances by 22 Japanese speakers with physiological variation.
It captures the lateral view of the speech production process including tongue and pharyngeal movements.
Each video consists of sequential MRIs and aligned audio files.
The data were recorded with 14 frame-per-second (FPS) or 27 FPS.
Each video starts with the resting position, which is in non-speaking states, and ends with the resting position.

Japanese dataset is suitable for our study because it is distinguished primarily based on tongue position and one of the most common vowel inventories~\citep {kubozonoHandbookJapanesePhonetics2015}.

\subsection{Tonguescape}

\paragraph{VowelVideo}
We extracted 120 silent real-time MRI videos from the rtMRIDB where the five basic isolated vowels (/a/, /i/, /u/, /e/, /o/) were pronounced. We split 5 samples (one for each vowel) as training data, another 5 samples as validation data, and the remaining 110 samples as test data. Since the tongue position during vowel articulation is not significantly affected by the pronunciation of preceding consonants, we also curated 1,653 videos where a vowel follows a consonant, such as in /ka/ and /na/, as additional training data, totaling 1,658 videos.
The test and validation data consist of videos for isolated vowel pronunciations.
The videos capture the entire articulation process, starting from a non-speaking state, progressing through the articulation, and returning to a non-speaking state.
Each video is around 1 or 2 seconds long and consists of about 14 or 27 frames per second.

\paragraph{VowelImage}
As shown in Figure~\ref{fig:dataset_example}, we manually selected one of the most characteristic and representative frames from the video of the five basic isolated vowels in the VowelVideo dataset. This allows us to separate the process of selecting representative moments from the video and estimating vowels from the characteristics of vowel images. 


\paragraph{VowelImageWithGuide}
Some studies show that adding markers as guides in images improves the performance for some tasks~\citep{shtedritski2023does}.
Inspired by this, we added guide markers to all images in the VowelImage dataset to facilitate the identification of tongue position within the oral cavity, similar to human perception, as shown in Figure~\ref{fig:dataset_example}. We drew ellipses as simple guides within the oral cavity in the images to encompass the palate, the body of the tongue during low vowel articulation, the tip of the tongue during front vowel articulation, and the root of the tongue during back vowel articulation. These guides were automatically drawn at similar coordinates across all images in the dataset and were manually checked.

\subsection{Dataset Difficulty}
Comparing the difficulty levels of VowelVideo and VowelImage is challenging.
While VowelImage allows for vowel prediction by observing and interpreting tongue position at a moment of articulation, it lacks the relative tongue movement information available in VowelVideo.

\begin{figure}[t]
    \centering
    \includegraphics[width=\columnwidth]{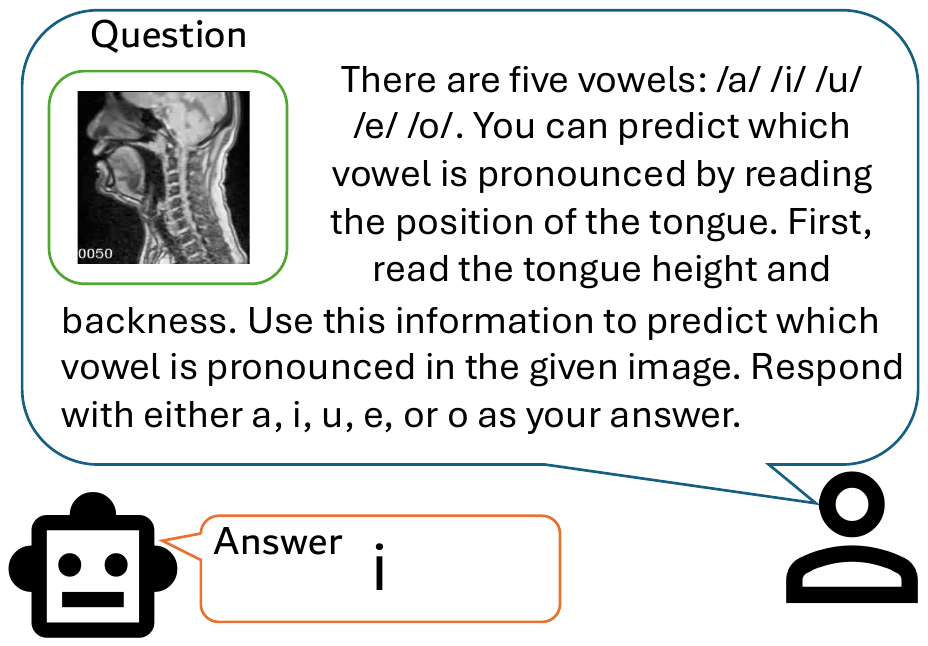}
    \caption{Example of the Instruction. We instruct LMs to predict the vowels from an inputted video/image. We evaluate whether the predicted vowels matched the vowels corresponding to the data used for an input. We specify the vowels (/a/, /i/, /u/, /e/, and /o/). Our prompt is designed as \citeposs{hu-levy-2023-prompting}.}
    \label{fig:experiment_example}
\end{figure}

\begin{figure*}[t]
    \centering
    \includegraphics[width=\textwidth]{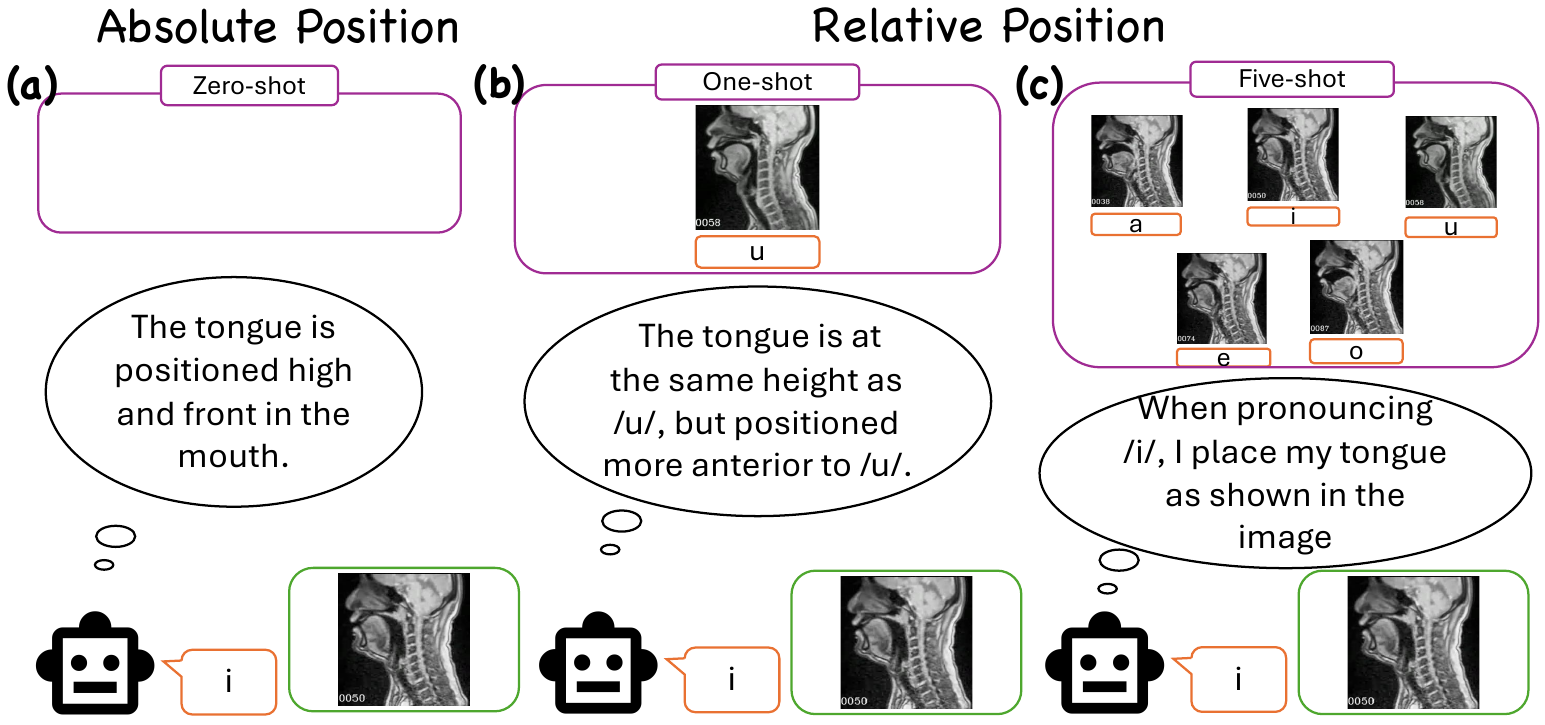}
    \caption{An overview of our methods. The ``Absolute Position'' (a) is an actual tongue position in an oral cavity. The ``Relative Position'' (b, c) is a tongue position at the moment of pronouncing a vowel regarding the position of other vowels.}
    \label{fig:overview_probing}
\end{figure*}

\section{Experiment}

\subsection{Experimental Settings}

We investigate the ability of LMs to predict vowels from input video/images using our dataset. Figure~\ref{fig:experiment_example} shows the prompt and example of our experiments\footnote{We used one or two NVIDIA RTX A6000 or NVIDIA A100 or four NVIDIA GeForce RTX 3090.}.
As baselines, we employed CLIP~\citep{radford2021learningtransferablevisualmodels} and fine-tuned CLIP (FT) with the VowelImage training set. CLIP is an image encoder model, and we used it to encode MRI images and classify them into five vowels.
The human baselines were created by a linguist for all datasets. We use accuracy as an evaluation metric.

\paragraph{VowelVideo}
We used Gemini 1.5 Pro~\citep{geminiteamGeminiUnlockingMultimodal2024}, GPT-4o~\citep{gpt4o}, LLaVA-NeXT-Interleave~\citep{li2024llavanext-interleave}, Phi-3.5-vision-instruct, Qwen2-VL-Instruct~\citep{Qwen2VL}, and VideoLLaMA2~\citep{damonlpsg2024videollama2}. We also fine-tuned VideoLLaMA2 (FT) using the VowelVideo training set.
For detailed experimental settings, please refer to Appendix~\ref{sec:detail-experimetal-settings-vowelvideo}.

\paragraph{VowelImage and VowelImageWithGuide}
We used Gemini 1.5 Pro~\citep{geminiteamGeminiUnlockingMultimodal2024}, GPT-4o~\citep{gpt4o}, LLaVA-NeXT-Interleave~\citep{li2024llavanext-interleave}, Phi-3.5-vision-instruct, VideoLLaMA2~\citep{damonlpsg2024videollama2}, LLaVA-Med~\citep{li2023llavamed}, Qwen-VL-Chat~\citep{Qwen-VL}, and Qwen2-VL-Instruct~\citep{Qwen2VL}. We also fine-tuned Qwen-VL-Chat (FT) using the VowelImage training set.
We conducted evaluations in 0-shot, 1-shot, and 5-shot settings. 
Note that we did not apply the 1-shot and 5-shot settings to VieoLLaMA2 and LLaVA-Med because they are not suitable for providing a question-answer example with an image.
For detailed experimental settings, please refer to Appendix~\ref{sec:detail-experimetal-settings-vowelimage}.


\begin{figure*}[!t]
    \centering
    \includegraphics[width=\textwidth]{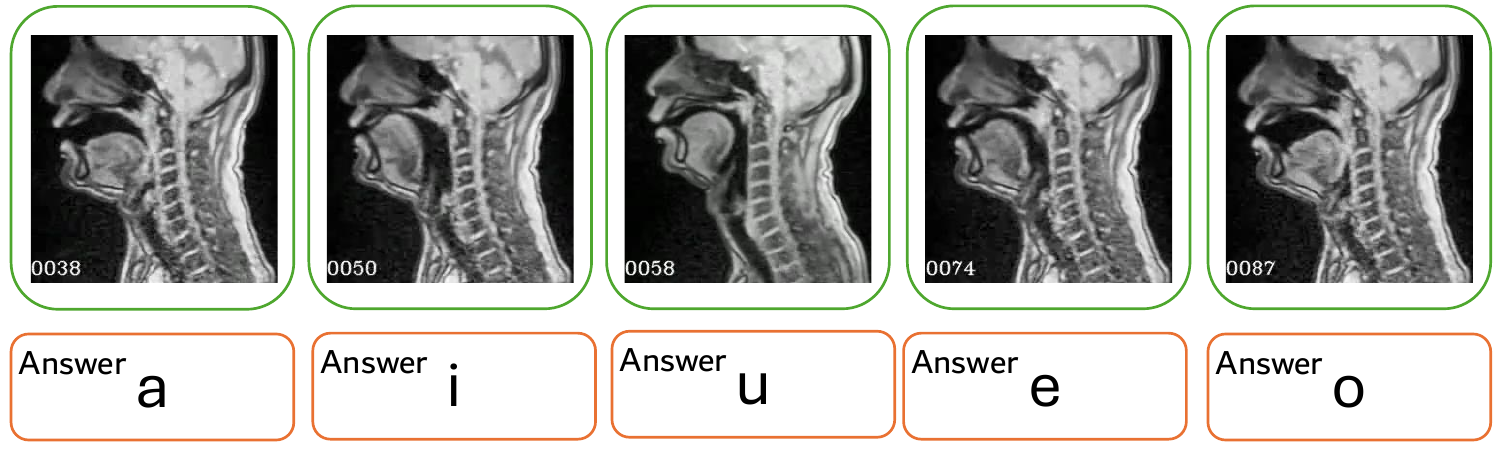}
    \caption{Five-shot examples of VowelImage.}
    \label{fig:five_shot_example}
\end{figure*}

\subsection{Few-shot Examples as Relative Positions of the Tongue: Bridging Linguistics}
\label{sec:few-shot-example-as-relative-position}

We divide our research question into those concerning \emph{absolute position} and those concerning \emph{relative position}.
The differences between these two positions are shown in Figure~\ref{fig:overview_probing}.
This is based on the fact that vowels can be distinguished by considering the relative position of the tongue~\citep{jonesEverymansEnglishPronouncing1917, knightCambridgeHandbookPhonetics2021a}.
``Absolute position'' of Figure~\ref{fig:overview_probing}(a) means an actual tongue position in an oral cavity when pronouncing a vowel.
``Relative position'' of Figure~\ref{fig:overview_probing}(b, c) is a tongue position in an oral cavity when pronouncing a vowel regarding a position when pronouncing another vowel.
There are two types of ``relative position'' in this context: relative position in terms of comparison with another vowel and relative position arising from physiological variations between speakers.
Dividing it in such a way allows us to gain a more detailed understanding of the capabilities of LMs.
Understanding absolute positions means they know the relationship between tongue positions and vowels and can read tongue positions in an image and associate that with a specific vowel.
Understanding relative positions means they understand the relationship between each vowel and can apply the knowledge of phonetics to predict vowels.
We can evaluate the capability of recognizing relative positions by using few-shot examples.

\paragraph{Relative Position in One-shot Example}
An understanding of the relationship between relative tongue position and vowels is evaluated by examining the one-shot performance of LMs.
Speakers can specify vowels from an MRI by referencing their own pronunciation experience and knowledge of tongue positions during speech, including information learned from MRIs and illustrations.
However, LMs appear to have little experience and information about tongue positions in the oral cavity.
Providing reference information about the tongue position of a vowel through a one-shot example is one of the approaches to bridge this gap, as shown in Figure~\ref{fig:overview_probing}(b).
If there is an image that indicates a vowel, the LMs can determine the tongue position from the given image using it.

\paragraph{Relative Position in Five-shot Examples}
The capability of understanding the relationship between tongue positions and vowels is evaluated by examining the few-shot performance of LMs or the performance of fine-tuned models.
The five-shot examples are shown in Figure~\ref{fig:five_shot_example}.
The number of examples is the same number of types of vowels used in this study.
If there are more than five images that contain each of the five vowels, they can determine which vowel the given image is closest to as in Figure~\ref{fig:overview_probing}(c).
This approach enables us to simulate the process by which human speakers predict vowels.

\paragraph{Relative Positions in Video}
We can observe relative positions in VowelVideo by looking at the tongue movements as mentioned in Section~\ref{sec:dataset}.
LMs handle a video as sequences of frames in the video.
In contrast to a few-shot example, this allows us to determine tongue positions relatively at the moment of articulation within an utterance.

\begin{table*}[t]
    \centering
    \small
    \begin{tabular}{lrrrrrrrrr}
       \toprule
         & VowelVideo &\multicolumn{3}{c}{VowelImage} & \multicolumn{3}{c}{VowelImageWithGuide} \\
        \cmidrule(lr){3-5} \cmidrule(l){6-8}
         & & zero-shot & one-shot & five-shot & zero-shot & one-shot & five-shot \\
       \midrule
       Random Choice & 20.00 & 20.00 & 20.00  & 20.00 & 20.00 & 20.00 & 20.00 \\
       CLIP & -- & 20.91 &  -- & -- & 24.55 & -- & -- \\
       CLIP (FT) & -- & 24.55 &  -- & -- & 29.09 & -- & -- \\
       Human & 71.55 & 61.82 &  -- & -- & 62.73 & -- & -- \\
       \midrule
       GPT-4o & 24.55 & \textbf{20.91} & \textbf{24.27}  & \textbf{37.27} & 12.73 & 20.18 & \textbf{40.00} \\
       Gemini 1.5 Pro & 18.18 & 16.36 &  21.27  & 34.55 & \textbf{20.00} & \textbf{22.73} & 34.55 \\
       LLaVA-NeXT-Interleave & 21.82 & 20.00 & 20.00 & 22.73 & \textbf{20.00} & 20.36 & 21.82 \\
       Phi-3.5-vision-instruct& 20.91 & 20.00 & 20.18  & 18.18 & 19.09 & 20.18 & 20.00 \\
       VideoLLaMA2 & 16.36 & 20.00 & -- & -- & 17.28 & -- & -- \\
       VideoLLaMA2 (FT) & \textbf{25.45} & 20.00 & -- & -- & \textbf{20.00} & -- & -- \\
       LLaVA-Med & -- & 11.82 &  -- & -- & 8.18 & -- & -- \\
       Qwen-VL-Chat & -- & 20.00 & 20.00  & 21.82 & \textbf{20.00} & 20.00 & 20.00 \\
       Qwen-VL-Chat (FT) & -- & 20.00 & 20.00 & 21.82 & \textbf{20.00} & 20.00 & 19.09 \\
       Qwen2-VL-7B-Instruct & 22.73 & 20.00 & 20.55  & 33.64 & 18.19 & 20.91 & 30.91 \\
       Qwen2-VL-72B-Instruct & 20.91 & 13.63 & 21.45  & 35.45 & 17.27 & 22.00 & \textbf{40.00} \\
       \bottomrule
    \end{tabular}
    \caption{Vowel prediction accuracy (\%) of different models in three settings. The accuracy is calculated by using the number of correct predictions and the number of test data.}
    \label{tab:result_accuracy_vowel}
\end{table*}

\subsection{Results}

\paragraph{VowelVideo}
Table~\ref{tab:result_accuracy_vowel} shows that the models struggle to predict vowels from videos correctly.
VideoLLaMA2 (FT) performs better than the original model, but the accuracy is still close to the chance rate.
This suggests that they are not well aligned for vowel information with video or sequential frames.

\paragraph{VowelImage}
As shown in Table~\ref{tab:result_accuracy_vowel}, the accuracy of each model is approximately 20\% in the zero-shot setting.
The accuracy improves in the one-shot and five-shot settings, particularly with Qwen2-VL-72B-Instruct.
This suggests that while some LMs can infer vowels by considering tongue positions relatively, they lack an understanding of absolute positions similar to that of linguists.
\paragraph{VowelImageWithGuide}
Table~\ref{tab:result_accuracy_vowel} shows that GPT-4o and Qwen2-VL-72B-Instruct performed much better in five-shot settings compared to VowelImage datasets.
However, other LMs still face challenges in predicting vowels from images with guides.
This suggests that LMs struggle to infer vowels when grounding in vision information.

\begin{figure*}[!t]
    \centering
    \includegraphics[width=\textwidth]{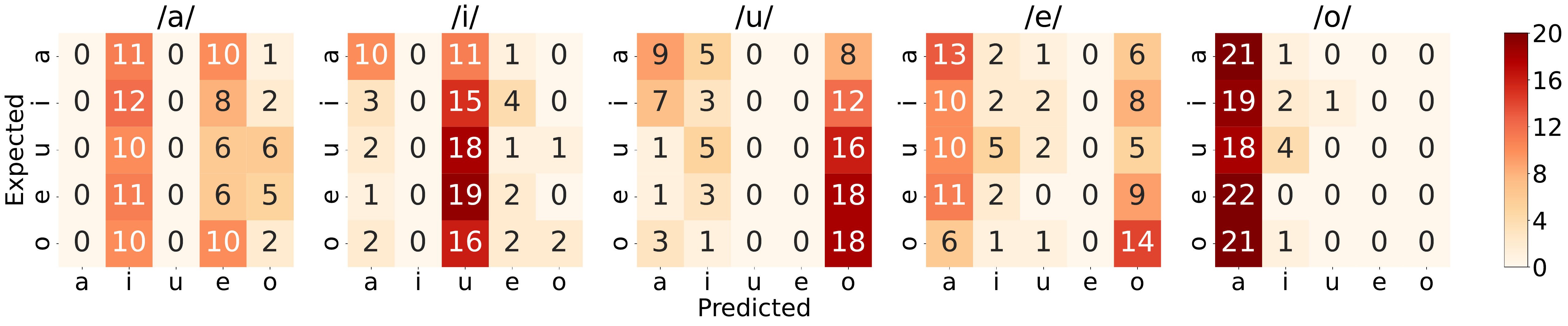}
    \caption{Confusion matrices of results using VowelImage and GPT-4o in the one-shot setting. Each matrix represents a confusion matrix obtained when the vowel images at the top of the figures are given as one-shot examples.}
    \label{fig:results_one_shot_gpt4o}
\end{figure*}

\section{Discussions and Analysis}

We will focus on important aspects in the following sections and defer more discussions to Appendix~\ref{sec:additional-discussions}.

\subsection{Comparison of each Dataset}
\paragraph{VowelVideo and VowelImage}
\label{sec:comparison_dataset}

Table~\ref{tab:result_accuracy_vowel} shows that it is easier for some models and a human to predict vowels from videos than from images in the zero-shot setting.
The video data contains not only the moment of articulation but also states before and after the articulation.
For example, we can look at relative positions during pronunciations.
This is a similar situation when using VowelImage in one-shot or five-shot settings.
Considering that some models have similar accuracies for VowelVideo and VowelImage in one-shot or five-shot settings, this information helps to predict vowels, although it could be noise.
The fact that some models and a human predict vowels better from videos than from images means that they use such information effectively.
Furthermore, these results demonstrate that these models can consider relative positions.

\paragraph{VowelImage and VowelImageWithGuide}
In Table~\ref{tab:result_accuracy_vowel}, the guideline results in poor performance in both zero-shot and one-shot settings when using GPT-4o, VideoLLaMA2, LLaVA-Med, and Qwen2-VL-7B-Instruct while the performances improve or remain unchanged when using the other models.
In the five-shot setting, it also has the same effect when using LLaVA-NeXT-Interleave, Qwen-VL-Chat, Qwen-VL-Chat (FT), and Qwen2-VL-7B-Instruct.
This suggests that the guideline is probably noise for some models and is a helpful guide for others.

\subsection{CLIP, LM and Tongue Positions}
\label{sec:clip_lm_tongue_positions}
VowelImage and VowelImageWithGuide in Table~\ref{tab:result_accuracy_vowel}, it is evident that most models in the zero-shot setting perform worse than CLIP.
However, in the five-shot setting, some models outperform the fine-tuned CLIP.
The underperformance of LMs compared to CLIP in the zero-shot setting suggests that LMs have difficulty predicting vowels based on absolute positions and understanding the association between vowels and tongue positioning in their training techniques.
In contrast, the five-shot enhancement implies that they can consider relative positions.
An analysis of the performance relative to CLIP reveals the ability of LMs to understand absolute positions and relative positions.

\subsection{Case Study: Analysis of the Results}
\label{sec:case_study_analysis}

We have analyzed the results of all LMs and highlighted two: a proprietary LM (GPT-4o) and an open LM (Qwen2-VL-72B-Instruct). The results of the other models can be found in Appendix~\ref{sec:additional-discussions}.

\subsubsection{Results of GPT-4o}
\label{sec:detail_gpt4o}

%

\paragraph{Zero-shot Setting}
The zero-shot setting results suggest that it is challenging for the model to read absolute tongue 
positions and predict vowels.

\paragraph{One-shot Setting}
As shown in Figure~\ref{fig:results_one_shot_gpt4o}, the one-shot setting results show that the given vowel was not reproduced in the output regardless of which vowel (/a/, /i/, /u/, /e/, or /o/) was provided as an example.
This suggests two key points: understanding of tongue position variability and interpretation of one-shot setting accuracy.
The model does not seem to comprehend that there is some degree of freedom in tongue positioning during articulation.
If it did, it would likely estimate the one-shot vowel to be the one with a tongue position closest to the provided image.
This effectively reduces the task from a five-choice problem to a four-choice problem.
Thus, the improvement in accuracy can be attributed to the reduction in choices rather than the model's understanding of relative tongue positions.
This result suggests that it struggles to recognize relative tongue positions given only one image.

\begin{figure}[t]
    \centering
    \includegraphics[width=\columnwidth]{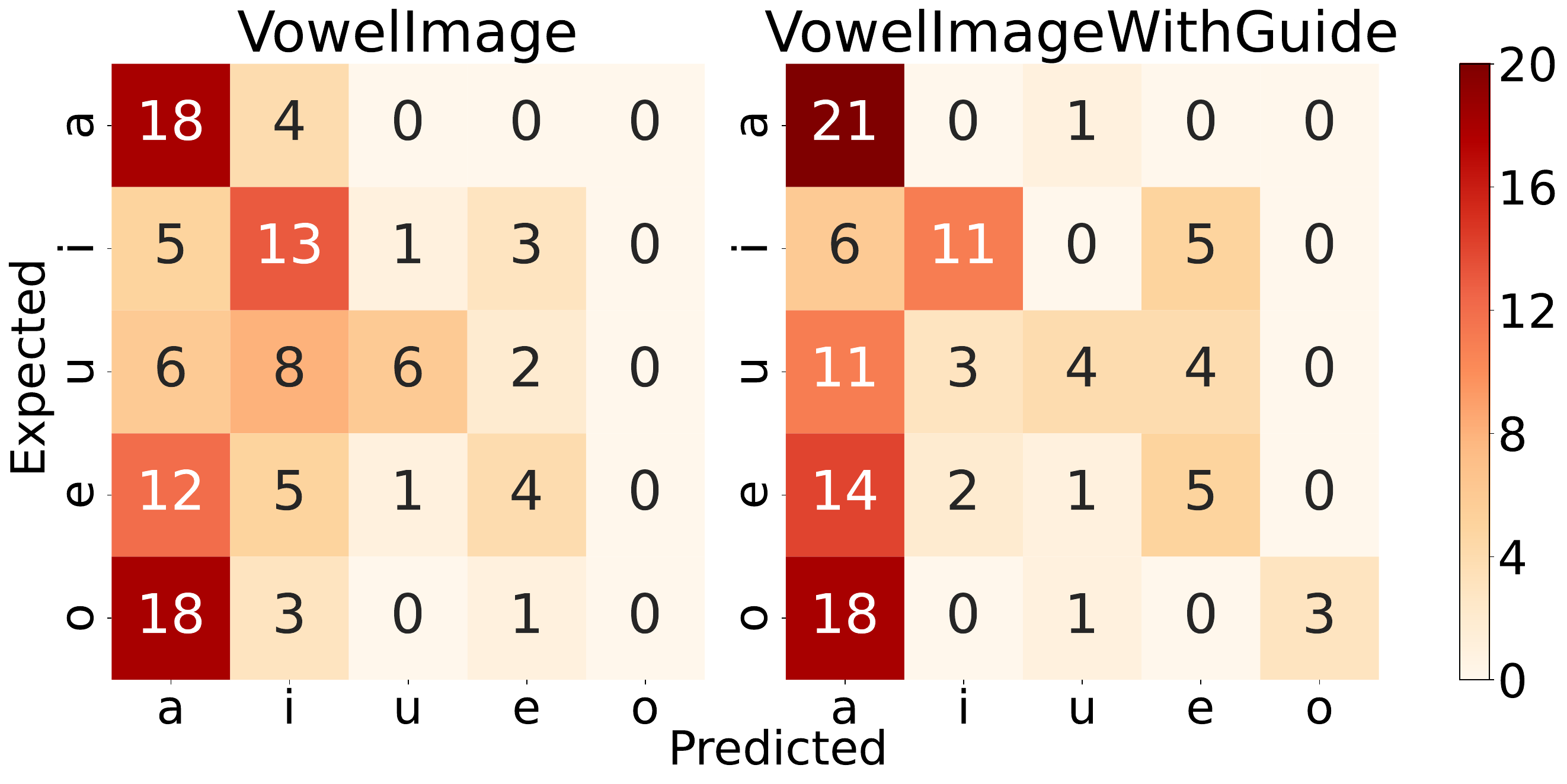}
    \caption{Confusion matrices of results of VowelImage and VowelImageWithGuide in GPT-4o in the five-shot setting.}
    \label{fig:result_gpt4o_five_shot_all}
\end{figure}

\paragraph{Five-shot Setting}
Figure~\ref{fig:result_gpt4o_five_shot_all} illustrates that the five-shot setting results show that the output is mainly /a / or /i /.
The model tended to output the high vowel /i/ when input images that indicate higher vowels (/i/, /u/) were provided, while it frequently outputs the low vowel /a/ for others.
The model seems to be able to recognize tongue height. 

\subsubsection{Results of Qwen2-VL-72B-Instruct}

\begin{figure*}[!t]
    \centering
    \includegraphics[width=\textwidth]{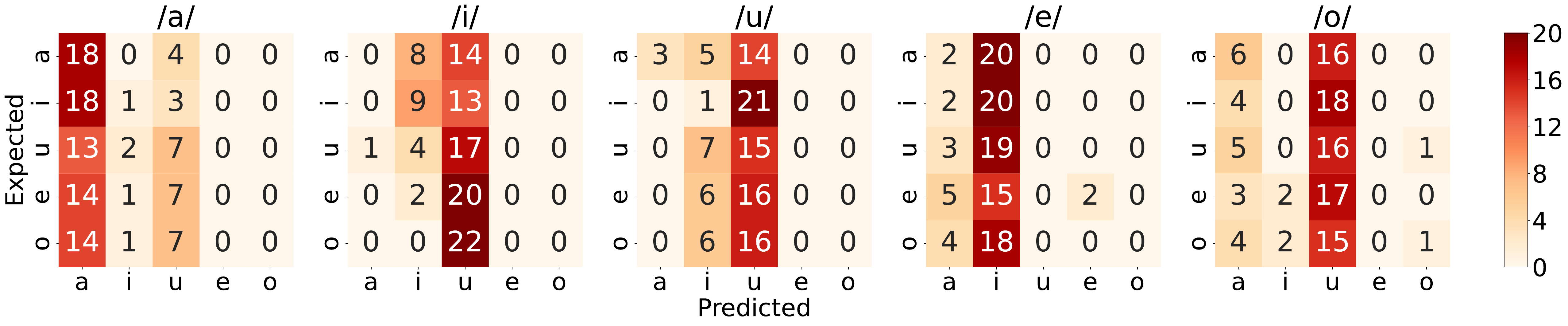}
    \caption{Confusion matrices of results using VowelImage and Qwen2-VL-72B-Instruct in the one-shot setting.}
    \label{fig:results_one_shot_qwen2vl72b}
\end{figure*}

\paragraph{Zero-shot Setting}
Table~\ref{tab:result_accuracy_vowel} indicates that this model performs worse than the baseline and random choice.
One of the reasons is that it did not predict any vowel for some images.
Although this leads to poor performance, it means that the model predicted vowels considering the given information.
However, even given these facts, the model struggles to predict vowels.

\paragraph{One-shot Setting}
In contrast to the results of GPT-4o shown in Figure~\ref{fig:results_one_shot_gpt4o}, in most cases, this model predicted the vowel given as a one-shot example as shown in Figure~\ref{fig:results_one_shot_qwen2vl72b}.
However, the model provided a vowel that has the same backness property when given /e/ and /o/ as a one-shot example.
This reveals that the model considers the positions of the tongue.

\begin{figure}[t]
    \centering
    \includegraphics[width=\columnwidth]{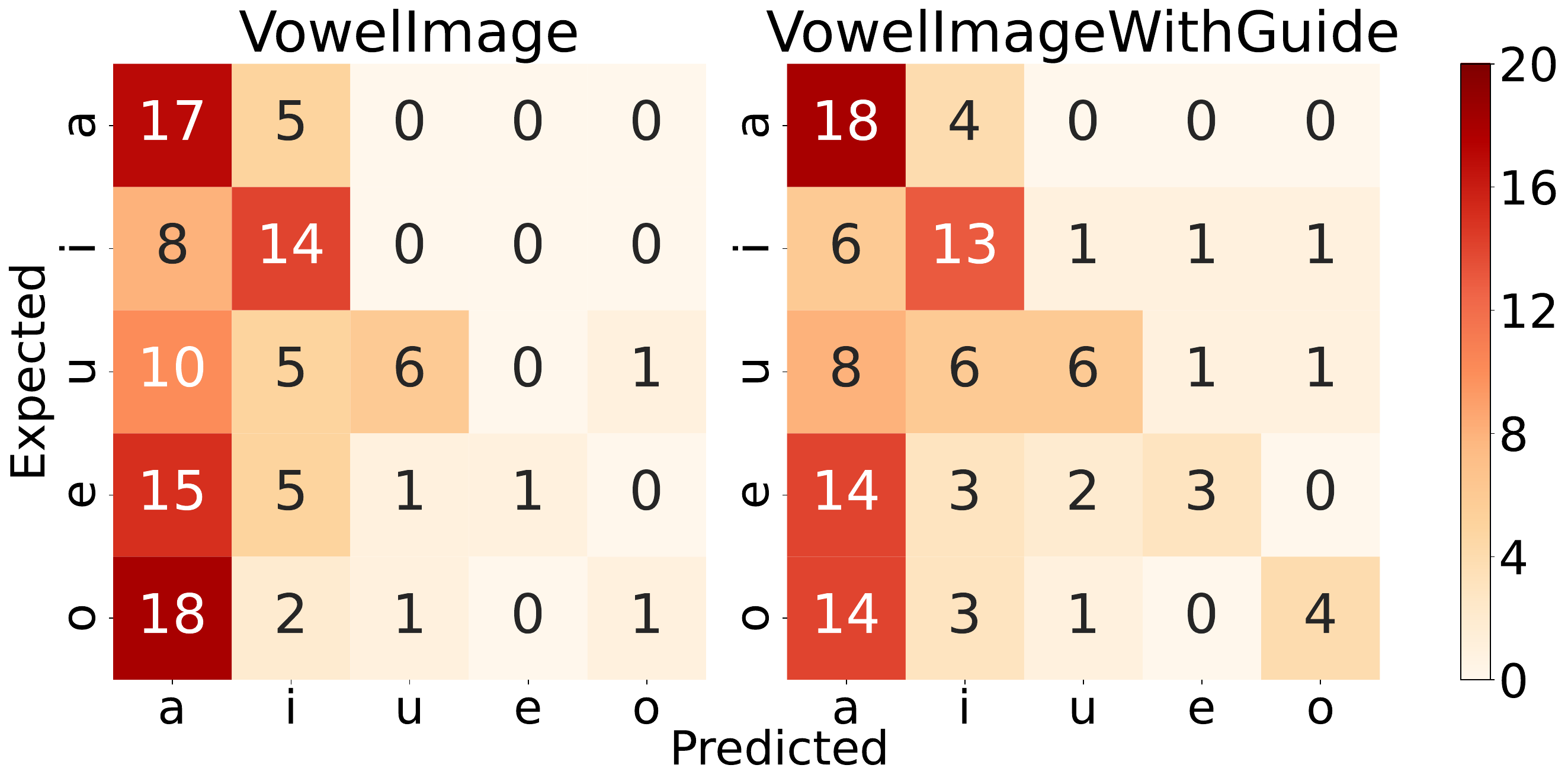}
    \caption{Confusion matrices of results of VowelImage and VowelImageWithGuide in Qwen2-VL-72b-Instruct in the five-shot setting.}
    \label{fig:result_qwen2vl72b_five_shot_all}
\end{figure}

\paragraph{Five-shot Setting}
Figure~\ref{fig:result_qwen2vl72b_five_shot_all} illustrates that most of the predictions are vowels /a/, /i/, and /u/.
In numerous instances, the model identified the vowel as /e/ and /o/ as /a/. This misclassification could be attributed to the fact that both /e/ and /o/, like /a/, are categorized as non-high vowels.

\begin{table*}[!t]
    \centering
    \small
    \setlength{\tabcolsep}{3.5pt}
    \begin{tabular}{lcccccc}
       \toprule
         & \multicolumn{3}{c}{VowelImage} & \multicolumn{3}{c}{VowelImageWithGuide} \\
        \cmidrule(lr){2-4} \cmidrule(l){5-7}
         & Height and Backness & Height & Backness & Height and Backness & Height & Backness \\
        \midrule
        Random Choice & 16.67 & 33.33 & 50.00 & 16.67 & 33.33 & 50.00 \\
        \midrule
        GPT-4o (directly) & 12.73 & 30.00 & 52.73 & 24.55 & 43.64 & 66.36 \\
        GPT-4o (vowel) & 37.27 & 46.36 & 50.90 & 40.00 & 42.73 & 47.27 \\
        Qwen2-VL-72B-Instruct (directly) & 20.91 & 46.36 & 43.64 & 21.82 & 46.36 & 44.55 \\
        Qwen2-VL-72B-Instruct (vowel) & 35.45 & 40.00 & 67.27 & 40.00 & 46.36 & 65.45 \\
        \bottomrule
    \end{tabular}
    \caption{Tongue height and backness prediction accuracies (\%) of GPT-4o and Qwen2-VL-72B-Instruct in a five-shot setting. The column named ``Height and Backness'' shows the accuracy that both ``Height'' and ``Backness'' are predicted correctly. GPT-4o (vowel) and Qwen2-VL-72B-Instruct (vowel) mean height and backness converted from predicted vowels in the five-shot setting. For example, they are ``high'' and ``front'' if /i/ is predicted. The definitions of the tongue position for each vowel are in Section~\ref{sec:backgroung_related_works}.}\label{tab:result_image_height_backness_gpt4o_five_shot}
\end{table*}

\subsection{Analysis of Fine-tuning Failuer}
We have fine-tuned VideoLLaMA2 and Qwen-VL-Chat, but they show limited improvement or decreased performance.
The limited improvement is caused because the training dataset is small, although the loss decreased for each model.
The decreased performance could be caused because this model outputs one of two vowels after fine-tuning while outputting the same vowel before fine-tuning. The fine-tuning adds variation to the output of this model, which may have resulted in lower performance for VowelImageWithGuide.

\subsection{Do LMs Consider Tongue Position?}
\label{sec:do_lms_consider_tongue_position}
In the one-shot setting, GPT-4o tended to infer mid or low vowels for most images of mid vowels while Qwen2-VL-72b-Instruct inferred high vowels as illustrated in Figures~\ref{fig:results_one_shot_gpt4o} and~\ref{fig:results_one_shot_qwen2vl72b}.
Figures~\ref{fig:result_gpt4o_five_shot_all} and~\ref{fig:result_qwen2vl72b_five_shot_all} illustrate that, in the five-shot setting, both GPT-4o and Qwen2-VL-72b-Instruct inferred /a/ for most images of mid vowels.
These findings clearly demonstrate that the models consistently predict vowels when analyzing images of mid vowels /e/ and /o/ in each setting.

\subsection{Do LMs Detect Tongue Positions?}
\label{sec:lms_detect_tongue_positions}
Table~\ref{tab:result_image_height_backness_gpt4o_five_shot} shows the results of predicting tongue height and backness both directly from images and by converting the estimated vowels to tongue positions, using GPT-4o and Qwen2-VL-72B-Instruct in the five-shot setting.
These results show that while the model is capable of predicting vowels from some images, it still struggles to predict tongue positions accurately.
These results indicate that LMs can consider tongue positions relatively.

\section{Conclusion: Do LMs Associate Tongue Positions with Vowel Articulation?}
We investigated the capability of LMs to understand the relationship between tongue positions and vowels to address our research question.

As shown in Table~\ref{tab:prediction_vowel_tonge}, LMs appear to understand the relationship between the positions of the tongue and the vowels.
However, the results of the zero-shot setting reveal that they have difficulty determining the tongue positions from a given image, as mentioned in Section~\ref{sec:clip_lm_tongue_positions}.
The results of fine-tuned models also demonstrate the challenge of learning the relationship through objective observation.
It can be posited that LMs appear to lack comprehension of tongue positions, as can be deduced from empirical knowledge and objective observation.

We have also discussed in Sections~\ref{sec:comparison_dataset},~\ref{sec:clip_lm_tongue_positions}, and~\ref{sec:do_lms_consider_tongue_position} that some LMs can consider relative positions to predict vowels.
These findings reveal that the LMs predict vowels with respect to other pairs of tongue positions and vowels while it is challenging to predict unseen vowels as shown in the results of the one-shot setting.
We can conclude that LMs, to a certain extent, associate tongue positions with the concept of vowel articulation.

In conclusion, our findings indicate whether LMs can comprehend the vowel articulation that linguists have long sought to decode.
While LMs faced challenges in our experiments, the performance of LMs can be improved like that of linguists when given examples.
We not only hope these findings will apply to large-scale linguistic analysis, speech synthesis, and educational fields but also wish for further research of languages.

\section{Limitations}

\paragraph{Multi-modal Language Models}
While our experiments were conducted using a limited set of LMs, there are few models capable of processing videos or multiple images simultaneously.
Given this context, our research can be considered comprehensive and a reasonable selection within the current state.

\paragraph{Generality of Dataset}
We investigated the performance of LMs using the Japanese five-vowel system, which is relatively easy to predict due to their distinct features.
As humans universally possess the same speech organs and share the capacity for pronunciation, we can generalize findings from Japanese data to human pronunciation more broadly.
Although there are many languages that have more or fewer than five vowel phonemes, we chose the five-vowel system for the following reasons: (1) a system with a small number of vowels would be easier to predict correctly, (2) classification of vowels in a larger vowel inventory solely based on MRI images could be partly an ambiguous task even for humans, and (3) the five-vowel system is one of the most common vowel inventories, and this size of a vowel inventory accounts for more than half of languages~\citep{wals-vowel-inventories}.
Although there are languages with a more complex vowel inventory than Japanese, the fact that models struggle even with relatively simple Japanese vowels suggests that using Japanese data as a first step is also a reasonable approach.
We intend to address languages with more complex vowel systems once the current challenges have been resolved.

\paragraph{Dataset Publicity}
The source dataset ``Real-time MRI Articulatory Movement Database - Version 1 (rtMRIDB)'' is licensed for research purposes only and does not allow sharing of derivatives or adaptations.

\paragraph{Performance improving}
We evaluated the performance of LMs using prompts.
There is room for improving the performance by some methods, e.g., chain-of-thought.
However, we aim to introduce a novel task and the baselines.
Consequently, incorporating strategies to enhance the efficacy of this task lies outside the boundaries of our study.
On the other hand, our findings that suggest one method, specifically few-shot prompting, outperformed others, indicate its potential effectiveness for this task.

\section{Ethical Considerations}

\paragraph{License of Source Dataset}
In this study, we have used the Real-time MRI Articulatory Movement Database (rtMRIDB)~\citep{maekawaRealtimeMRIArticulatory2022} to create our dataset.
This dataset is licensed only for research purposes.
Since we have been permitted to use this dataset by the providing institution, there are no licensing issues.

\paragraph{Identifying Infomation and Offensive Content}
Our datasets are created from the Real-time MRI Articulatory Movement Database.
We have confirmed that the original dataset does not contain any personally identifying information or offensive content, thus our dataset also does not contain them.
We have also confirmed that no inappropriate content is included in our dataset.


\paragraph{Use of AI Assistants}
In this study, we have used GitHub Copilot as an AI assistant for coding support.

\section{Acknowledgement}
In this study, we used ``Real-time MRI Articulatory Movement Database - Version 1 (rtMRIDB)'' developed by National Institute for Japanese Language and Linguistics and provided by Speech Resources Consortium, National Institute of Informatics.
We thank Chihiro Taguchi and the anonymous reviewers for their valuable comments and suggestions. This work was supported by JSPS KAKENHI Grant Number JP23H03458.


\bibliography{anthology,custom}

\begin{thebibliography}{32}
\providecommand{\natexlab}[1]{#1}

\bibitem[{Abeysinghe et~al.(2022)Abeysinghe, James, Watson, and Marattukalam}]{abeysingheVisualisingModelTraining2022}
Binu~Nisal Abeysinghe, Jesin James, Catherine Watson, and Felix Marattukalam. 2022.
\newblock \href {https://doi.org/10.21437/Interspeech.2022-264} {Visualising {{Model Training}} via {{Vowel Space}} for {{Text-To-Speech Systems}}}.
\newblock In \emph{Proc. {{Interspeech}} 2022}, pages 511--515.

\bibitem[{Bai et~al.(2023)Bai, Bai, Yang, Wang, Tan, Wang, Lin, Zhou, and Zhou}]{Qwen-VL}
Jinze Bai, Shuai Bai, Shusheng Yang, Shijie Wang, Sinan Tan, Peng Wang, Junyang Lin, Chang Zhou, and Jingren Zhou. 2023.
\newblock \href {https://doi.org/10.48550/arXiv.2308.12966} {{Qwen-VL}: A versatile vision-language model for understanding, localization, text reading, and beyond}.
\newblock \emph{arXiv preprint arXiv:2308.12966}.

\bibitem[{Bybee(2015)}]{Bybee_2015}
Joan Bybee. 2015.
\newblock \emph{Language Change}.
\newblock Cambridge Textbooks in Linguistics. Cambridge University Press.

\bibitem[{Cao et~al.(2022)Cao, Han, Long, Xu, and Poon}]{caoUnderstandingAttentionVisionLanguage2022}
Feiqi Cao, Soyeon~Caren Han, Siqu Long, Changwei Xu, and Josiah Poon. 2022.
\newblock Understanding {{Attention}} for {{Vision-and-Language Tasks}}.
\newblock In \emph{Proceedings of the 29th {{International Conference}} on {{Computational Linguistics}}}, pages 3438--3453, Gyeongju, Republic of Korea. International Committee on Computational Linguistics.

\bibitem[{Cheng et~al.(2024)Cheng, Leng, Zhang, Xin, Li, Chen, Zhu, Zhang, Luo, Zhao, and Bing}]{damonlpsg2024videollama2}
Zesen Cheng, Sicong Leng, Hang Zhang, Yifei Xin, Xin Li, Guanzheng Chen, Yongxin Zhu, Wenqi Zhang, Ziyang Luo, Deli Zhao, and Lidong Bing. 2024.
\newblock \href {https://arxiv.org/abs/2406.07476} {{VideoLLaMA 2}: Advancing spatial-temporal modeling and audio understanding in video-llms}.
\newblock \emph{arXiv preprint arXiv:2406.07476}.

\bibitem[{Cotterell and Eisner(2017)}]{cotterell-eisner-2017-probabilistic}
Ryan Cotterell and Jason Eisner. 2017.
\newblock \href {https://doi.org/10.18653/v1/P17-1109} {Probabilistic typology: Deep generative models of vowel inventories}.
\newblock In \emph{Proceedings of the 55th Annual Meeting of the Association for Computational Linguistics (Volume 1: Long Papers)}, pages 1182--1192, Vancouver, Canada. Association for Computational Linguistics.

\bibitem[{Cotterell and Eisner(2018)}]{cotterell-eisner-2018-deep}
Ryan Cotterell and Jason Eisner. 2018.
\newblock \href {https://doi.org/10.18653/v1/N18-1004} {A deep generative model of vowel formant typology}.
\newblock In \emph{Proceedings of the 2018 Conference of the North {A}merican Chapter of the Association for Computational Linguistics: Human Language Technologies, Volume 1 (Long Papers)}, pages 37--46, New Orleans, Louisiana. Association for Computational Linguistics.

\bibitem[{{Gemini Team}(2024)}]{geminiteamGeminiUnlockingMultimodal2024}
{Gemini Team}. 2024.
\newblock \href {https://doi.org/10.48550/arXiv.2403.05530} {Gemini 1.5: {{Unlocking}} multimodal understanding across millions of tokens of context}.
\newblock \emph{Preprint}, arXiv:2403.05530.

\bibitem[{Hayashi et~al.(2024)Hayashi, Sakai, Kamigaito, Hayashi, and Watanabe}]{hayashi-etal-2024-towards}
Kazuki Hayashi, Yusuke Sakai, Hidetaka Kamigaito, Katsuhiko Hayashi, and Taro Watanabe. 2024.
\newblock \href {https://aclanthology.org/2024.acl-short.65} {Towards artwork explanation in large-scale vision language models}.
\newblock In \emph{Proceedings of the 62nd Annual Meeting of the Association for Computational Linguistics (Volume 2: Short Papers)}, pages 705--729, Bangkok, Thailand. Association for Computational Linguistics.

\bibitem[{Hu et~al.(2022)Hu, Shen, Wallis, Allen-Zhu, Li, Wang, Wang, and Chen}]{hu2022lora}
Edward~J Hu, Yelong Shen, Phillip Wallis, Zeyuan Allen-Zhu, Yuanzhi Li, Shean Wang, Lu~Wang, and Weizhu Chen. 2022.
\newblock \href {https://openreview.net/forum?id=nZeVKeeFYf9} {Lo{RA}: Low-rank adaptation of large language models}.
\newblock In \emph{International Conference on Learning Representations}.

\bibitem[{Hu and Levy(2023)}]{hu-levy-2023-prompting}
Jennifer Hu and Roger Levy. 2023.
\newblock \href {https://doi.org/10.18653/v1/2023.emnlp-main.306} {Prompting is not a substitute for probability measurements in large language models}.
\newblock In \emph{Proceedings of the 2023 Conference on Empirical Methods in Natural Language Processing}, pages 5040--5060, Singapore. Association for Computational Linguistics.

\bibitem[{{International PhoneticAssociation}(1999)}]{associationHandbookInternationalPhonetic1999}
{International PhoneticAssociation}. 1999.
\newblock \emph{Handbook of the {{International Phonetic Association}}: {{A Guide}} to the {{Use}} of the {{International Phonetic Alphabet}}}.
\newblock Cambridge University Press.

\bibitem[{Jones(1917)}]{jonesEverymansEnglishPronouncing1917}
Daniel Jones. 1917.
\newblock \emph{{{English Pronouncing Dictionary}}}.
\newblock Dent, London.

\bibitem[{Joseph(2018)}]{joseph-2018-historical}
Andrew~Jonathan Joseph. 2018.
\newblock \emph{The {{Historical Phonology}} of {{Manchu Dialects}}}.
\newblock Ph.D. thesis, Cornell University, Ann Arbor, United States.

\bibitem[{Kawaharazuka et~al.(2024)Kawaharazuka, Matsushima, Gambardella, Guo, Paxton, and Zeng}]{kawaharazukaRealWorldRobotApplications2024}
Kento Kawaharazuka, Tatsuya Matsushima, Andrew Gambardella, Jiaxian Guo, Chris Paxton, and Andy Zeng. 2024.
\newblock \href {https://doi.org/10.48550/arXiv.2402.05741} {Real-{{World Robot Applications}} of {{Foundation Models}}: {{A Review}}}.
\newblock \emph{Preprint}, arXiv:2402.05741.

\bibitem[{Knight and Setter(2021)}]{knightCambridgeHandbookPhonetics2021a}
Rachael-Anne Knight and Jane Setter, editors. 2021.
\newblock \href {https://doi.org/10.1017/9781108644198} {\emph{The {{Cambridge Handbook}} of {{Phonetics}}}}.
\newblock Cambridge {{Handbooks}} in {{Language}} and {{Linguistics}}. Cambridge University Press, Cambridge.

\bibitem[{Ko(2012)}]{ko-2012-tongue}
Seong Ko. 2012.
\newblock \emph{Tongue {{Root Harmony And Vowel Contrast In Northeast Asian Languages}}}.
\newblock Ph.D. thesis, Cornell University.

\bibitem[{Kubozono(2015)}]{kubozonoHandbookJapanesePhonetics2015}
Haruo Kubozono, editor. 2015.
\newblock \href {https://doi.org/doi:10.1515/9781614511984} {\emph{Handbook of Japanese Phonetics and Phonology}}.
\newblock De Gruyter Mouton, Berlin, M{\"u}nchen, Boston.

\bibitem[{Li et~al.(2023)Li, Wong, Zhang, Usuyama, Liu, Yang, Naumann, Poon, and Gao}]{li2023llavamed}
Chunyuan Li, Cliff Wong, Sheng Zhang, Naoto Usuyama, Haotian Liu, Jianwei Yang, Tristan Naumann, Hoifung Poon, and Jianfeng Gao. 2023.
\newblock {{LLaVA-Med}}: {{Training}} a large language-and-vision assistant for biomedicine in one day.
\newblock In \emph{Advances in Neural Information Processing Systems}, volume~36, pages 28541--28564. Curran Associates, Inc.

\bibitem[{Li et~al.(2024)Li, Zhang, Zhang, Zhang, Li, Li, Ma, and Li}]{li2024llavanext-interleave}
Feng Li, Renrui Zhang, Hao Zhang, Yuanhan Zhang, Bo~Li, Wei Li, Zejun Ma, and Chunyuan Li. 2024.
\newblock \href {https://doi.org/10.48550/arXiv.2407.07895} {{{LLaVA-NeXT-Interleave}}: {{Tackling Multi-image}}, {{Video}}, and {{3D}} in {{Large Multimodal Models}}}.
\newblock \emph{Preprint}, arXiv:2407.07895.

\bibitem[{Maddieson(2013)}]{wals-vowel-inventories}
Ian Maddieson. 2013.
\newblock \href {https://doi.org/10.5281/zenodo.7385533} {Vowel quality inventories (v2020.3)}.
\newblock In Matthew~S. Dryer and Martin Haspelmath, editors, \emph{The World Atlas of Language Structures Online}. Zenodo.

\bibitem[{Maekawa(2022)}]{maekawaRealtimeMRIArticulatory2022}
Kikuo Maekawa. 2022.
\newblock Real-time {{MRI Articulatory Movement Database}} ({{rtMRIDB}}).

\bibitem[{OpenAI(2024)}]{gpt4o}
OpenAI. 2024.
\newblock \href {https://arxiv.org/abs/2410.21276} {Gpt-4o system card}.
\newblock \emph{Preprint}, arXiv:2410.21276.

\bibitem[{Pal and Sankarasubbu(2024)}]{pal-sankarasubbu-2024-gemini}
Ankit Pal and Malaikannan Sankarasubbu. 2024.
\newblock \href {https://doi.org/10.18653/v1/2024.clinicalnlp-1.3} {Gemini goes to {M}ed school: Exploring the capabilities of multimodal large language models on medical challenge problems {\&} hallucinations}.
\newblock In \emph{Proceedings of the 6th Clinical Natural Language Processing Workshop}, pages 21--46, Mexico City, Mexico. Association for Computational Linguistics.

\bibitem[{Radford et~al.(2021)Radford, Kim, Hallacy, Ramesh, Goh, Agarwal, Sastry, Askell, Mishkin, Clark, Krueger, and Sutskever}]{radford2021learningtransferablevisualmodels}
Alec Radford, Jong~Wook Kim, Chris Hallacy, Aditya Ramesh, Gabriel Goh, Sandhini Agarwal, Girish Sastry, Amanda Askell, Pamela Mishkin, Jack Clark, Gretchen Krueger, and Ilya Sutskever. 2021.
\newblock \href {https://arxiv.org/abs/2103.00020} {Learning transferable visual models from natural language supervision}.
\newblock \emph{Preprint}, arXiv:2103.00020.

\bibitem[{Shtedritski et~al.(2023)Shtedritski, Rupprecht, and Vedaldi}]{shtedritski2023does}
Aleksandar Shtedritski, Christian Rupprecht, and Andrea Vedaldi. 2023.
\newblock \href {https://doi.org/10.1109/ICCV51070.2023.01101} {What does clip know about a red circle? visual prompt engineering for vlms}.
\newblock In \emph{2023 IEEE/CVF International Conference on Computer Vision (ICCV)}, pages 11953--11963, Los Alamitos, CA, USA. IEEE Computer Society.

\bibitem[{Steuer et~al.(2023)Steuer, List, Abdullah, and Klakow}]{steuer-etal-2023-information}
Julius Steuer, Johann-Mattis List, Badr~M. Abdullah, and Dietrich Klakow. 2023.
\newblock \href {https://doi.org/10.18653/v1/2023.sigtyp-1.10} {Information-theoretic characterization of vowel harmony: A cross-linguistic study on word lists}.
\newblock In \emph{Proceedings of the 5th Workshop on Research in Computational Linguistic Typology and Multilingual NLP}, pages 96--109, Dubrovnik, Croatia. Association for Computational Linguistics.

\bibitem[{Wang(2020)}]{WangManZhouYuXianDaiFangYanniokeruMuYinDiaoHe2020}
Haibo Wang. 2020.
\newblock \href {http://hdl.handle.net/2115/77601} {Vowel harmony in the modern dialects of manchu}.
\newblock \emph{Northern Language Studies}, 10:135--156.

\bibitem[{Wang et~al.(2024)Wang, Bai, Tan, Wang, Fan, Bai, Chen, Liu, Wang, Ge, Fan, Dang, Du, Ren, Men, Liu, Zhou, Zhou, and Lin}]{Qwen2VL}
Peng Wang, Shuai Bai, Sinan Tan, Shijie Wang, Zhihao Fan, Jinze Bai, Keqin Chen, Xuejing Liu, Jialin Wang, Wenbin Ge, Yang Fan, Kai Dang, Mengfei Du, Xuancheng Ren, Rui Men, Dayiheng Liu, Chang Zhou, Jingren Zhou, and Junyang Lin. 2024.
\newblock {Qwen2-VL}: Enhancing vision-language model's perception of the world at any resolution.
\newblock \emph{arXiv preprint arXiv:2409.12191}.

\bibitem[{Wu et~al.(2023)Wu, Li, Lu, Zhang, Lian, Black, Goldstein, Watanabe, and Anumanchipalli}]{wuDeepSpeechSynthesis2023}
Peter Wu, Tingle Li, Yijing Lu, Yubin Zhang, Jiachen Lian, Alan~W. Black, Louis Goldstein, Shinji Watanabe, and Gopala~K. Anumanchipalli. 2023.
\newblock \href {https://doi.org/10.21437/Interspeech.2023-2316} {Deep {{Speech Synthesis}} from {{MRI-Based Articulatory Representations}}}.
\newblock In \emph{Proc. {{Interspeech}} 2023}, pages 5132--5136.

\bibitem[{Yan et~al.(2024)Yan, He, Yue, and Wang}]{yanWorseRandomEmbarrassingly2024}
Qianqi Yan, Xuehai He, Xiang Yue, and Xin~Eric Wang. 2024.
\newblock \href {https://openreview.net/forum?id=ZJZ6rkFxNO} {Worse than random? an embarrassingly simple probing evaluation of large multimodal models in medical {VQA}}.
\newblock In \emph{GenAI for Health: Potential, Trust and Policy Compliance}.

\bibitem[{Zhou et~al.(2023)Zhou, Lai, Yeong, Mouratidis, and Jiang}]{zhou-etal-2023-rome}
Kankan Zhou, Eason Lai, Wei Bin~Au Yeong, Kyriakos Mouratidis, and Jing Jiang. 2023.
\newblock \href {https://doi.org/10.18653/v1/2023.findings-emnlp.683} {{ROME}: Evaluating pre-trained vision-language models on reasoning beyond visual common sense}.
\newblock In \emph{Findings of the Association for Computational Linguistics: EMNLP 2023}, pages 10185--10197, Singapore. Association for Computational Linguistics.

\end{thebibliography}

\appendix

\section{Dataset Details}
\label{sec:detail-rtmridb}

\paragraph{Dataset Statictics}
Tabale~\ref{tab:dataset_stats} shows the dataset statics.
Each of the datasets, VowelImage and VowelImageWithGuide, consists of 120 videos or images.
From these, we selected 5 samples (one for each vowel) as training data, another 5 samples as development data, and the remaining 110 samples as test data.
The training and development data are from a single participant each, and these participants' data are not included in the test data.
VowelVideo consists of 1773 videos.
The development and test data are equivalent to those used in VowelImage and VowelImageWithGuide.
The other videos are the test dataset.
Each test data contains five vowels equally.

\section{Details of the Experimental Settings}
\label{sec:detail-experimetal-settings}

\subsection{Prediction from Real-time MRI (video)}
\label{sec:detail-experimetal-settings-vowelvideo}
We used GPT-4o, Gemini 1.5 Pro~\citep{geminiteamGeminiUnlockingMultimodal2024}, LLaVA-NeXT-Interleave~\citep{li2024llavanext-interleave}, Phi-3.5-vision-instruct, VideoLLaMA2~\citep{damonlpsg2024videollama2}, and Qwen2-VL-Instruct~\citep{Qwen2VL}. We also fine-tuned VideoLLaMA2 (FT) using VowelVideo training data with LoRA~\citep{hu2022lora}.
Hyperparameters of fine-tuning are in Table~\ref{tab:experiment_param_vieollama2_qwenvl}.
The details of these models are in Table~\ref{tab:model_detail}.

We make both videos that are recorded at 14 frame-per-second (FPS) and videos recorded at 27 FPS 14 times slower to input to Gemini 1.5 Pro because a video is sampled at 1 FPS~\citep{geminiteamGeminiUnlockingMultimodal2024}.
Only 16 frames are used in each video in VideoLLaMA2 due to its limitation, while all are used in GPT-4o, LLaVA-NeXT-Interleave, and Phi-3.5-vision-instruct.
When using Qwen2-VL-Instruct, we handle the input video in 14 FPS regardless of its FPS.

We use only a vowel in the output as a predicted vowel for evaluations when we use fine-tuned VideoLLaMA2 if the output is a consonant-vowel pair.

\begin{table}[t]
    \centering
    \setlength{\tabcolsep}{3pt}
    \begin{tabular}{lrrrr}
        \toprule
         & train & dev & test & total \\
        \midrule
        VowelVideo & 1,658 & 5 & 110 & 1,773 \\
        VowelImage & 5 & 5 & 110 & 120 \\
        VowelImagewithGuide & 5 & 5 & 110 & 120 \\
        \bottomrule
    \end{tabular}
    \caption{Dataset statics. In VowelVideo, a vowel of the answer is considered to calculate.}
    \label{tab:dataset_stats}
\end{table}

\begin{table*}[t]
    \centering
    \small
    \begin{tabular}{ll}
        \toprule
        Model & Model ID \\
        \midrule
        CLIP & openai/clip-vit-large-patch14 \\
        GPT-4o & gpt-4o-2024-05-13 \\
        Gemini 1.5 Pro & models/gemini-1.5-pro-001 \\
        LLaVA-NeXT-Interleave & lmms-lab/llava-next-interleave-qwen-7b-dpo \\
        Phi-3.5-vision-instruct & microsoft/Phi-3.5-vision-instruct \\
        VideoLLaMA2 & DAMO-NLP-SG/VideoLLaMA2-7B \\
        Qwen-VL-Chat & Qwen/Qwen-VL-Chat \\
        Qwen-VL-7B-Instruct & Qwen/Qwen2-VL-7B-Instruct \\
        Qwen-VL-72B-Instruct & Qwen/Qwen2-VL-72B-Instruct-GPTQ-Int4 \\
        \bottomrule
    \end{tabular}
    \caption{The model details. Model ID is a Huggingface Repository ID or a code defined in OpenAI API and Gemini API.}
    \label{tab:model_detail}
\end{table*}

\begin{table*}[t]
    \centering
    \small
    \begin{tabular}{lrrr}
        \toprule
        Hyperparameter & CLIP & VideoLLaMA2 & Qwen-VL-Chat \\
        \cmidrule(r){1-1} \cmidrule(lr){2-2} \cmidrule(l){3-3} \cmidrule(l){4-4}
        batch size & 5 & 4 & 1 \\
        epoch & 40 & 1 & 40 \\
        learning rate & 1e-4 & 2e-5 & 1e-5 \\
        seed & 42 & 42 & 42 \\
        warmup ratio & 0.0 & 0.03 & 0.1 \\
        \bottomrule
    \end{tabular}
    \caption{Hyperparameters to fine-tune CLIP, VideoLLaMA2 and Qwen-VL-Chat}
    \label{tab:experiment_param_vieollama2_qwenvl}
\end{table*}

\subsection{Prediction from one MRI (image)}
\label{sec:detail-experimetal-settings-vowelimage}

We used GPT-4o, Gemini 1.5 Pro~\citep{geminiteamGeminiUnlockingMultimodal2024}, LLaVA-NeXT-Interleave~\citep{li2024llavanext-interleave}, Phi-3.5-vision-instruct, VideoLLaMA2~\citep{damonlpsg2024videollama2}, LLaVA-Med~\citep{li2023llavamed}, Qwen-VL-Chat~\citep{Qwen-VL}, and Qwen2-VL-Instruct~\citep{Qwen2VL}.
We used CLIP~\citep{radford2021learningtransferablevisualmodels} for the baseline.
We also fine-tuned Qwen-VL-Chat (FT) using VowelImage training data with LoRA~\citep{hu2022lora} and CLIP (FT) using VowelIamge training data.
The hyperparameters of fine-tuning are in Table~\ref{tab:experiment_param_vieollama2_qwenvl}.
The details of the models are in Table~\ref{tab:model_detail}.
Especially, GPT-4o is one of the most suitable models for our experiments because it performs well in clinical tasks overall although it struggles in several tasks, e.g., position description.
We converted images into videos when using VideoLLaMA2 because it accepts only video files.
We conducted experiments in three settings, zero-shot, one-shot, and five-shot, to see the capability of handling absolute positions and relative positions.
The reasons why we use few-shot examples to explore relative positions are explained in Section~\ref{sec:few-shot-example-as-relative-position}.

\paragraph{Zero-shot Setting}
Each image in the dataset is used as an input to LMs.
When we use GPT-4o and Gemini 1.5 Pro, we specify a JSON schema, such as \code{\{"vowel": str\}}, to output only the predicted vowel.
If the output is not only vowels, we extract the first vowel surrounded by `` '' or / / and consider it the predicted vowel.
We treat the model as having refused to answer the question if there is no vowel surrounded by them.

\paragraph{One-shot Setting}
One of the training data is selected as a one-shot example.
For each training data, we use all test data and evaluate the accuracy.
We give the models the one-shot example using a conversation template.
We use the conversation template defined in each model except Gemini 1.5 Pro.
In Gemini 1.5 Pro, we add an example in the prompt.
We do not apply this setting to VideoLLaMA2 and LLaVA-Med because they are not suitable for providing a question-answer example with an image.

\paragraph{Five-shot Setting}
All of the training data are used as five-shot examples such as in Figure~\ref{fig:five_shot_example}.
This means an image corresponding to each of the five vowels is used.
We give the models the five-shot examples using a conversation template.
We also use the conversation template defined in each model except Gemini 1.5 Pro.
In Gemini 1.5 Pro, we add examples in the prompt.
The order of the five-shot examples is the same in all models: /a/ image, /i/ image, /u/ image, /e/ image, and /o/ image.
We do not apply this setting to VideoLLaMA2 and LLaVA-Med because they are not suitable for providing question-answer examples with images.

\section{Additional Discussions}
\label{sec:additional-discussions}


\subsection{Analysis of the Results of GPT-4o (Detail)}
Tabel~\ref{tab:result_image_gpt4o_one_shot} and Table~\ref{tab:result_image_guide_gpt4o_one_shot} show the proportions of inferred vowels for each expected vowel when using GPT-4o.
The results demonstrate that the model infrequently classified the image as the vowel provided as a one-shot example, suggesting that the model considers the position similarity strictly.

\begin{table}[t]
    \centering
    \small
    \setlength{\tabcolsep}{4pt}
    \begin{tabular}{ccccccc}
    \toprule
     One-shot& Expected & \multicolumn{5}{c}{Predicted Vowel} \\
    exemple& Vowel & /a/ & /i/ & /u/ & /e/ & /o/ \\
    \cmidrule(r){1-1} \cmidrule(lr){2-2} \cmidrule(l){3-7}
    \multirow{5}{*}{/a/} & /a/ & \textbf{0.00} & 0.50 & 0.00 & 0.45 & 0.05 \\
    & /i/ & \textbf{0.00} & 0.55 & 0.00 & 0.36 & 0.09 \\
    & /u/ & \textbf{0.00} & 0.45 & 0.00 & 0.27 & 0.27 \\
    & /e/ & \textbf{0.00} & 0.50 & 0.00 & 0.27 & 0.23 \\
    & /o/ & \textbf{0.00} & 0.45 & 0.00 & 0.45 & 0.09 \\
    \midrule
    \multirow{5}{*}{/i/} & /a/ & 0.45 & \textbf{0.00} & 0.50 & 0.05 & 0.00 \\
    & /i/ & 0.14 & \textbf{0.00} & 0.68 & 0.18 & 0.00 \\
    & /u/ & 0.09 & \textbf{0.00} & 0.82 & 0.05 & 0.05 \\
    & /e/ & 0.05 & \textbf{0.00} & 0.86 & 0.09 & 0.00 \\
    & /o/ & 0.09 & \textbf{0.00} & 0.73 & 0.09 & 0.09 \\
    \midrule
    \multirow{5}{*}{/u/} & /a/ & 0.41 & 0.23 & \textbf{0.00} & 0.00 & 0.36 \\
    & /i/ & 0.32 & 0.14 & \textbf{0.00} & 0.00 & 0.55 \\
    & /u/ & 0.05 & 0.23 & \textbf{0.00} & 0.00 & 0.73 \\
    & /e/ & 0.05 & 0.14 & \textbf{0.00} & 0.00 & 0.82 \\
    & /o/ & 0.14 & 0.05 & \textbf{0.00} & 0.00 & 0.82 \\
    \midrule
    \multirow{5}{*}{/e/} & /a/ & 0.59 & 0.09 & 0.05 & \textbf{0.00} & 0.27 \\
    & /i/ & 0.45 & 0.09 & 0.09 & \textbf{0.00} & 0.36 \\
    & /u/ & 0.45 & 0.23 & 0.09 & \textbf{0.00} & 0.23 \\
    & /e/ & 0.50 & 0.09 & 0.00 & \textbf{0.00} & 0.41 \\
    & /o/ & 0.27 & 0.05 & 0.05 & \textbf{0.00} & 0.64 \\
    \midrule
    \multirow{5}{*}{/o/}& /a/ & 0.95 & 0.05 & 0.00 & 0.00 & \textbf{0.00} \\
    & /i/ & 0.86 & 0.09 & 0.05 & 0.00 & \textbf{0.00} \\
    & /u/ & 0.82 & 0.18 & 0.00 & 0.00 & \textbf{0.00} \\
    & /e/ & 1.00 & 0.00 & 0.00 & 0.00 & \textbf{0.00} \\
    & /o/ & 0.95 & 0.05 & 0.00 & 0.00 & \textbf{0.00} \\
    \bottomrule
    \end{tabular}
    \caption{The proportions of vowels output for each expected vowel using VowelImage and GPT-4o with the one-shot setting. Each table shows the results of each one-shot example. The bold numbers indicate the percentage of the vowel used as a one-shot example.}
    \label{tab:result_image_gpt4o_one_shot}
\end{table}

\begin{table}[t]
    \centering
    \small
    \setlength{\tabcolsep}{4pt}
    \begin{tabular}{ccccccc}
    \toprule
     One-shot& Expected & \multicolumn{5}{c}{Predicted Vowel} \\
    exemple& Vowel & /a/ & /i/ & /u/ & /e/ & /o/ \\
    \cmidrule(r){1-1} \cmidrule(lr){2-2} \cmidrule(l){3-7}
    \multirow{5}{*}{/a/} & /a/ & \textbf{0.00} & 0.00 & 0.00 & 0.91 & 0.09 \\
    & /i/ & \textbf{0.00} & 0.00 & 0.00 & 0.59 & 0.41 \\
    & /u/ & \textbf{0.00} & 0.09 & 0.00 & 0.55 & 0.36 \\
    & /e/ & \textbf{0.00} & 0.05 & 0.00 & 0.64 & 0.32 \\
    & /o/ & \textbf{0.00} & 0.14 & 0.00 & 0.55 & 0.32 \\
    \midrule
    \multirow{5}{*}{/i/} & /a/ & 0.00 & \textbf{0.00} & 0.14 & 0.86 & 0.00 \\
    & /i/ & 0.00 & \textbf{0.00} & 0.00 & 1.00 & 0.00 \\
    & /u/ & 0.00 & \textbf{0.00} & 0.05 & 0.95 & 0.05 \\
    & /e/ & 0.00 & \textbf{0.00} & 0.05 & 0.95 & 0.00 \\
    & /o/ & 0.05 & \textbf{0.00} & 0.23 & 0.68 & 0.05 \\
    \midrule
    \multirow{5}{*}{/u/} & /a/ & 0.00 & 0.00 & \textbf{0.00} & 0.05 & 0.95 \\
    & /i/ & 0.00 & 0.05 & \textbf{0.00} & 0.00 & 0.95 \\
    & /u/ & 0.00 & 0.09 & \textbf{0.00} & 0.09 & 0.82 \\
    & /e/ & 0.00 & 0.00 & \textbf{0.00} & 0.00 & 1.00 \\
    & /o/ & 0.00 & 0.00 & \textbf{0.00} & 0.05 & 0.95 \\
    \midrule
    \multirow{5}{*}{/e/} & /a/ & 0.05 & 0.27 & 0.00 & \textbf{0.00} & 0.68 \\
    & /i/ & 0.00 & 0.18 & 0.00 & \textbf{0.00} & 0.82 \\
    & /u/ & 0.00 & 0.23 & 0.09 & \textbf{0.00} & 0.77 \\
    & /e/ & 0.09 & 0.18 & 0.00 & \textbf{0.05} & 0.68 \\
    & /o/ & 0.00 & 0.14 & 0.00 & \textbf{0.00} & 0.86 \\
    \midrule
    \multirow{5}{*}{/o/}& /a/ & 0.32 & 0.00 & 0.05 & 0.64 & \textbf{0.00} \\
    & /i/ & 0.64 & 0.00 & 0.05 & 0.32 & \textbf{0.00} \\
    & /u/ & 0.45 & 0.00 & 0.00 & 0.55 & \textbf{0.00} \\
    & /e/ & 0.59 & 0.00 & 0.05 & 0.36 & \textbf{0.00} \\
    & /o/ & 0.36 & 0.00 & 0.05 & 0.59 & \textbf{0.00} \\
    \bottomrule
    \end{tabular}
    \caption{The propositions of predicted vowels for each expected vowel using VowelImageGuide and GPT-4o with the one-shot setting. Each table shows the results of each one-shot example. The bold numbers indicate the percentage of the vowel used as a one-shot example.}
    \label{tab:result_image_guide_gpt4o_one_shot}
\end{table}

\subsection{Analysis of the Results of Gemini 1.5 Pro}
\begin{figure}[!t]
    \centering
    \includegraphics[width=\columnwidth]{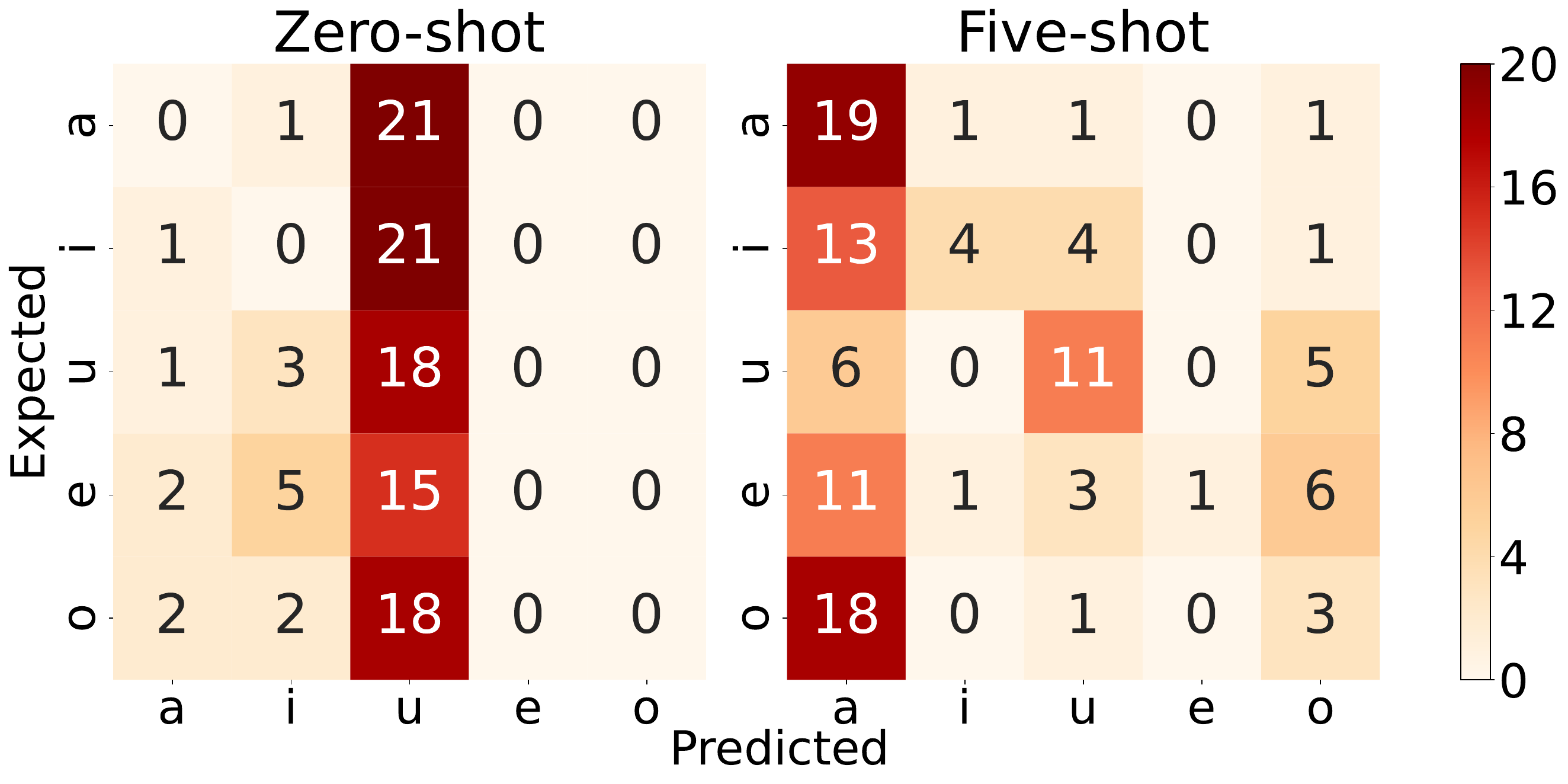}
    \caption{Confusion matrices of results using VowelImage in Gemini 1.5 Pro in the zero-shot and five-shot settings.}
    \label{fig:result_image_video_gemini}
\end{figure}

As shown in Table~\ref{tab:result_accuracy_vowel}, this model
and Figure~\ref{fig:result_image_video_gemini}, this model predicts vowels more correctly in the five-shot setting than in the zero-shot setting.
In the one-shot setting, the performance of VowelImage demonstrates an improvement compared to the zero-shot setting, with this model achieving the highest accuracy for VowelImageWithGuide. 
Nevertheless, the extent of this improvement is limited, and the overall performance remains comparable to that of random choice.

\subsection{Analysis of the Results of LLaVA-NeXT-Interleave}
\begin{figure}[!t]
    \centering
    \includegraphics[width=\columnwidth]{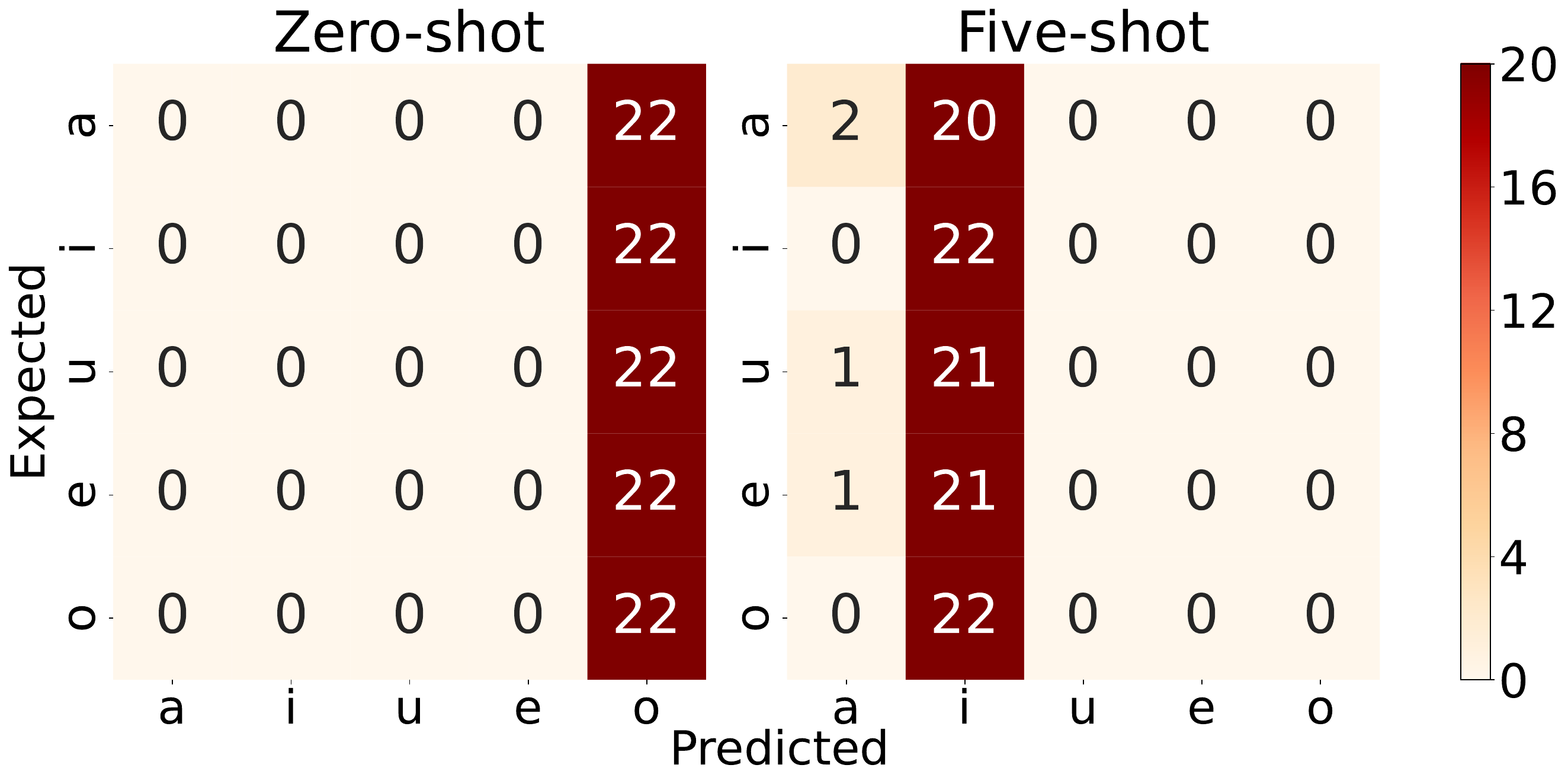}
    \caption{Confusion matrices of results using VowelImage and LLaVA-NeXT-Interleave in the zero-shot and five-shot settings.}\label{fig:result_image_llavanext_zero_five_shot}
\end{figure}

Table~\ref{tab:result_accuracy_vowel} shows that the accuracy of this model is improved by providing five-shot examples.
However, Fiugure~\ref{fig:result_image_llavanext_zero_five_shot} shows that it gave the same output for almost all inputs, and struggles to predict vowels even when provided several examples.

\subsection{Analysis of Results of Phi-3.5-vision-instruct}
\begin{figure}[!t]
    \centering
    \includegraphics[width=\columnwidth]{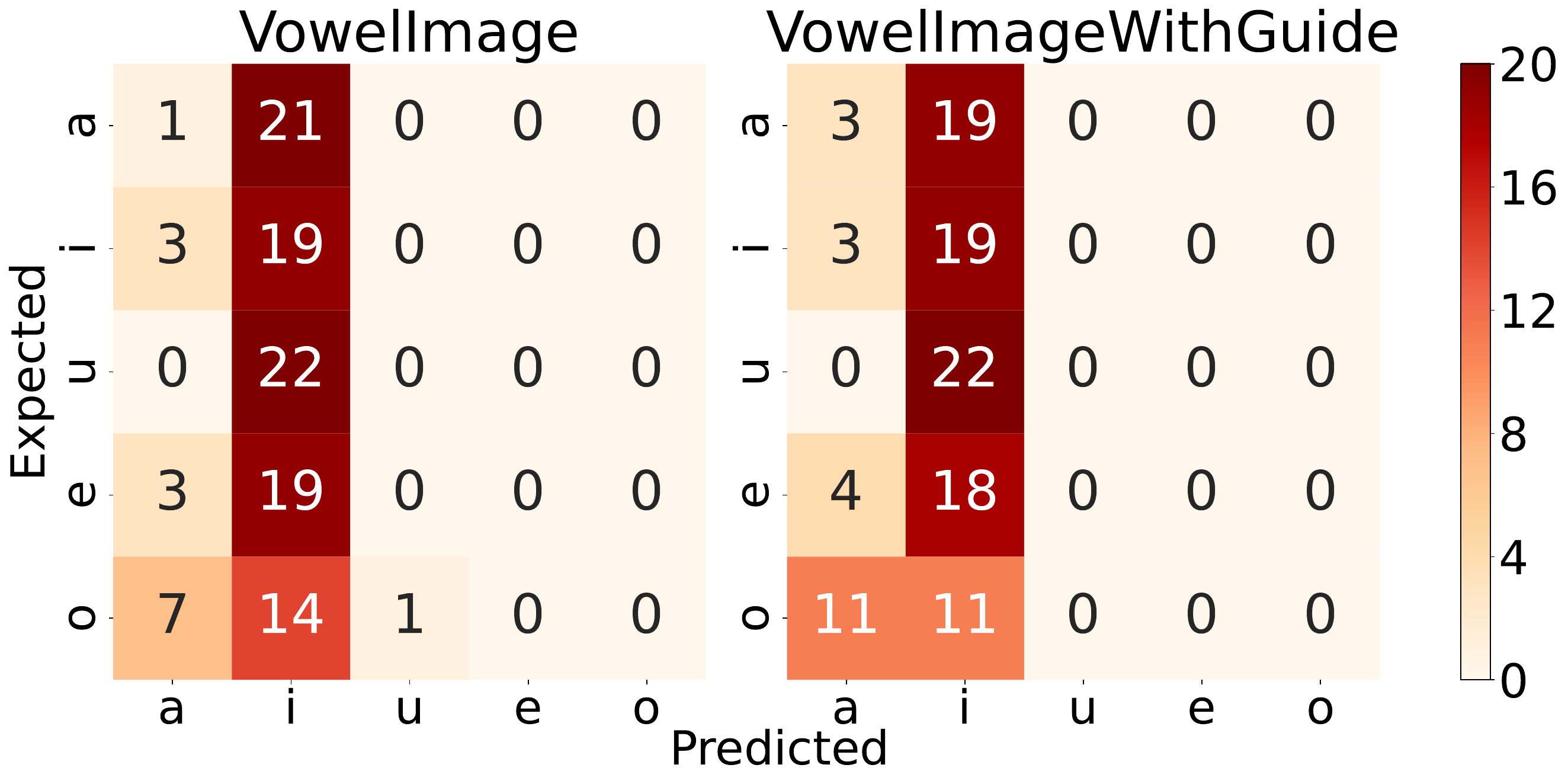}
    \caption{Confusion matrices of results of VowelImage and VowelImageWithGuide in Phi-3.5-vision-instruct in the five-shot setting.}
    \label{fig:result_image_phi35_five_shot}
\end{figure}

Table~\ref{tab:result_accuracy_vowel} shows that only this model performed worse when using VowelImageWithGuide than when using VowelImage.
Figure~\ref{fig:result_image_phi35_five_shot} illustrates that the two matrices show similar distributions and that the reason for the difference in performance is likely attributable to an error.
The performance in the five-shot setting for VowelImage is also decreased.
This indicates that this model struggles to compare multiple images.

\subsection{Analysis of the Results of VideoLLaMA2}
\begin{figure}[t]
     \centering
     \includegraphics[width=0.6\columnwidth]{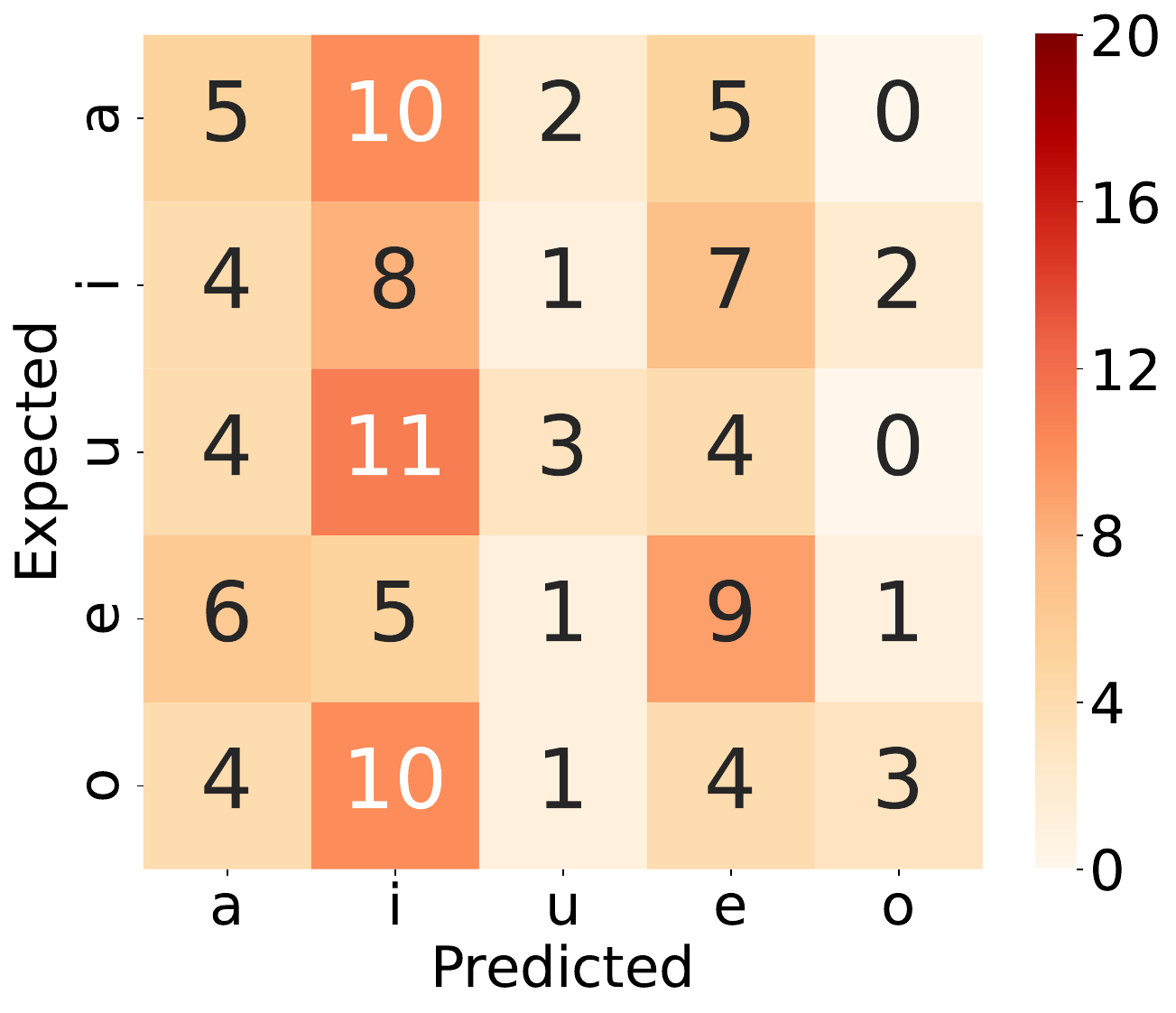}
     \caption{A Confusion matrix of results using VowelVideo and VideoLLaMA2 (FT).}
     \label{fig:result_video_videollama2}
 \end{figure}

Table~\ref{tab:result_accuracy_vowel} and Figure~\ref{fig:result_video_videollama2} show that the performance using VowelVideo was improved by fine-tuning.
However, this model still struggles to predict vowels from images.
The reason for this suggests that either the model was fine-tuned by a video recording of the entire utterance, or that this model may not be good at handling videos with only one frame.

\subsection{Analysis of the Results of LLaVA-Med}
\begin{figure}[!t]
    \centering
    \includegraphics[width=\columnwidth]{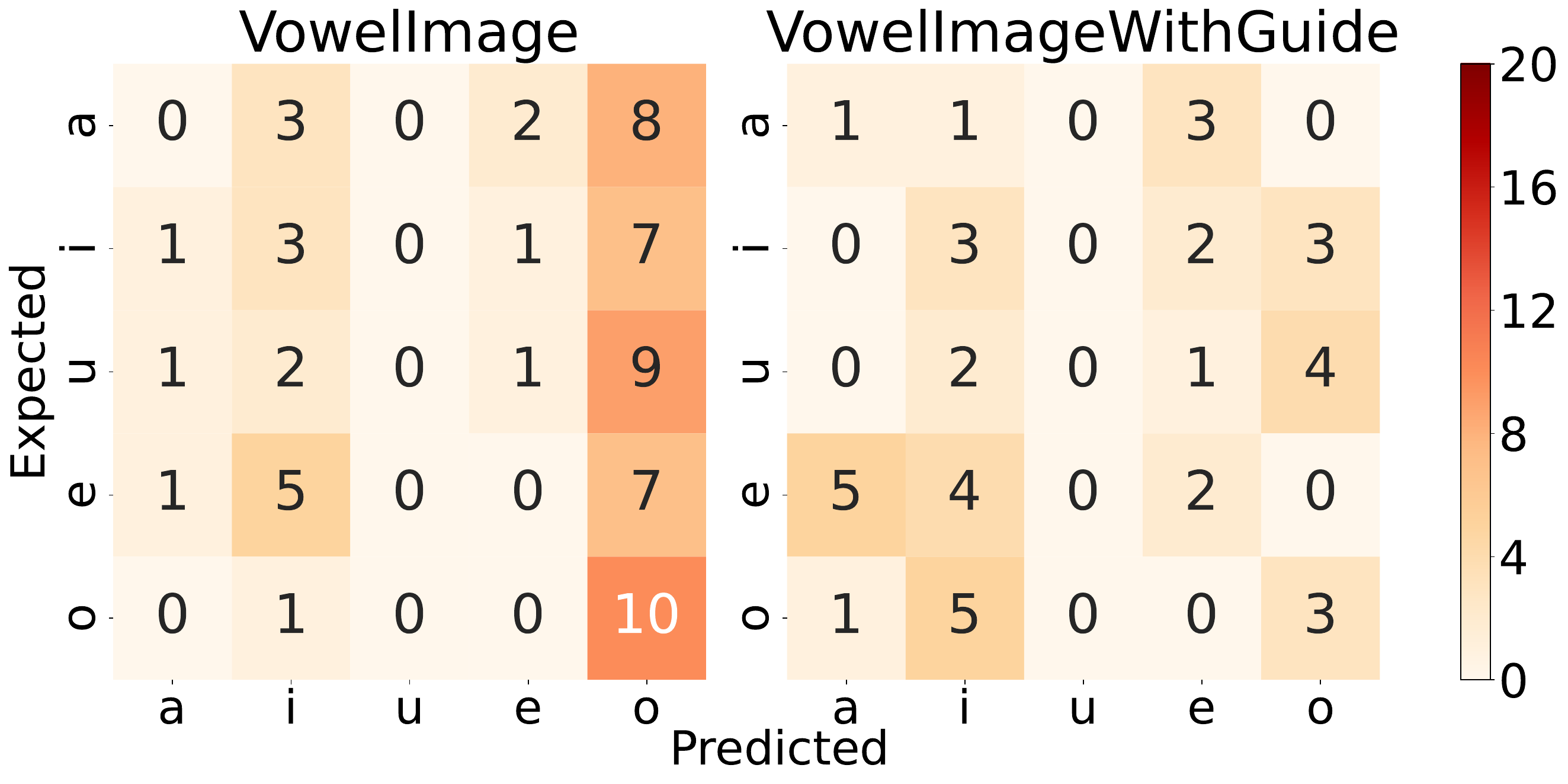}
    \caption{Confusion matrices of results using VowelImage and VowelImageWithGuide in LLaVA-Med in the zero-shot setting. The rejected outputs were removed from this heatmap.}
    \label{fig:result_llava_med}
\end{figure}

LLaVA-Med is trained with clinical data but struggles to predict vowels, as shown in Table~\ref{tab:result_accuracy_vowel} and Figure~\ref{fig:result_llava_med}.
One of the reasons for this low accuracy is that it rejected to answer.
For example, it rejected to answer for 48 samples when using VowelImage.
This means they tried to predict vowels for 62 samples and The percentage of correct answers to the number of attempted answers is 20.97 \%.
When using VowelImageWithGuide in the zero-shot setting, the predicted vowels are distributed compared with that in other models in the zero-shot setting.

\subsection{Analysis of the Results of Qwen-VL-Chat}

\begin{figure}[t]
    \centering
    \includegraphics[width=\columnwidth]{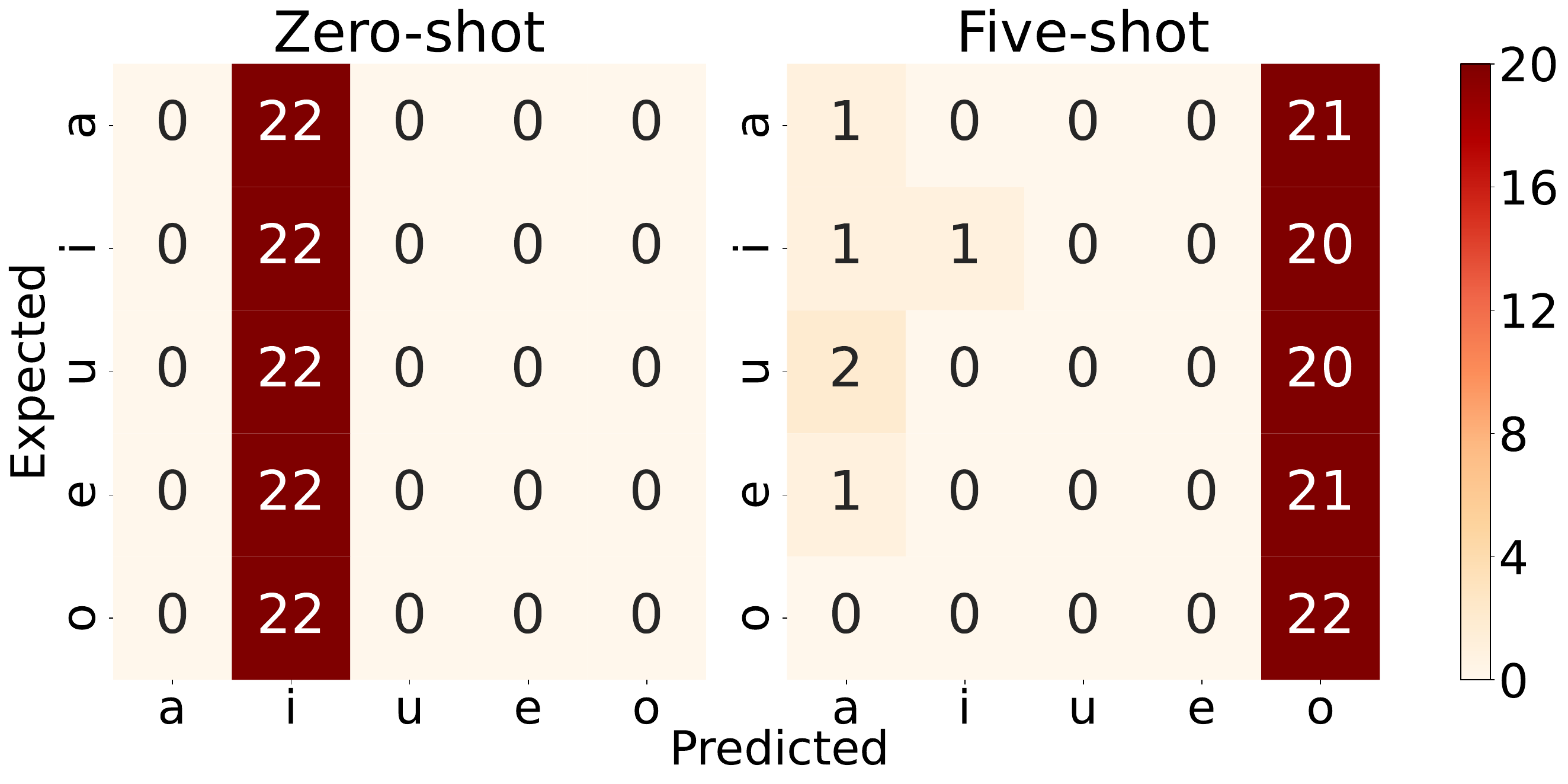}
    \caption{Confusion matrices of results using VowelImage and Qwen-VL-Chat in the zero-shot and five-shot settings.}
    \label{fig:result_image_qwenvl}
\end{figure}

Table~\ref{tab:result_accuracy_vowel} shows that the accuracies of Qwen-VL-Chat when using VowelImage and VowelImageWithGuide in both the zero-shot and one-shot settings are the same as the chance rate.
This is because they output the same vowel in those settings (see Figure~\ref{fig:result_image_qwenvl}).
The fine-tuned model, Qwen-VL-Chat (FT), also performed with similar accuracies while the loss decreased from 0.2144 to 0.1545.
One of the reasons why the performance of the fine-tuned model remained could be the small number of training data.

\vspace{0.78em}
\subsection{Analysis of Results of Qwen2-VL-7B-Instruct}

\begin{figure}[!]
    \centering
    \includegraphics[width=\columnwidth]{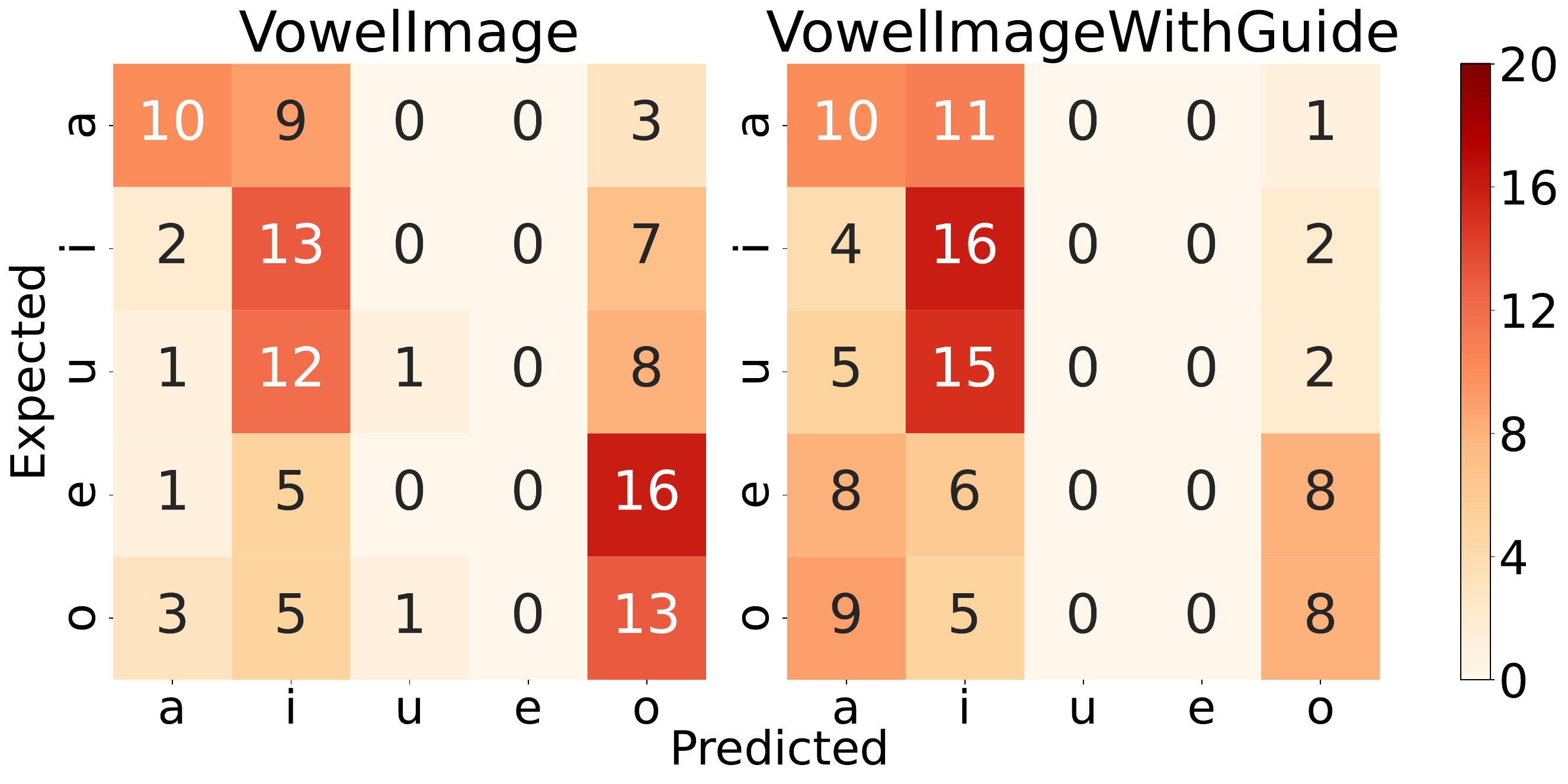}
    \caption{Confusion matrices of results of VowelImage and VowelImageWithGuide in Qwen2-VL-7B-Instruct in the five-shot setting.}
    \label{fig:result_image_qwen2vl7b_five_shot}
\end{figure}

Table~\ref{tab:result_accuracy_vowel} shows that the accuracy when using VowelImageWithGuide is worse than when using VowelImage.
When using VowelImageWithGuide, this model inferred the vowel as /the low-back vowel a/  lfor images that represent the mnon-high vowel /e/ or  /o/ in some instances, as shown in Figure~\ref{fig:result_image_qwen2vl7b_five_shot}, causing the accuracy to decline.
For the high vowels /i/ and /u/, the model predicted /i/.
The results demonstrate that this model can determine whether the provided image represents a high vowel or not both for VowelImage and VowelImageWithGuide.

\end{document}